\documentclass[10pt,journal,compsoc]{IEEEtran}
\usepackage{graphicx}
\usepackage{comment}
\usepackage{amsmath,amssymb} 
\usepackage{color}

\usepackage{epsfig}
\usepackage{booktabs}
\usepackage{multirow}
\usepackage[export]{adjustbox}
\usepackage{float}
\usepackage{dashrule}
\usepackage{arydshln}
\usepackage[utf8x]{inputenc}
\usepackage{subcaption}
\usepackage{url}
\usepackage{algorithm}
\usepackage{algorithmic}

\usepackage{xcolor}
\renewcommand\fbox{\fcolorbox{gray}{white}}
\newcommand{\etal}{et al.}

\ifCLASSOPTIONcompsoc
\usepackage[nocompress]{cite}
\else
\usepackage{cite}
\fi
\ifCLASSINFOpdf
\else
\fi
\begin{document}
%
\title{HoughNet: Integrating near and long-range evidence for visual detection}
%
%
%
%

\author{Nermin~Samet\textsuperscript{\textdagger}, Samet~Hicsonmez,
    and~Emre~Akbas
\thanks{ N. Samet and E. Akbas are  at the Dept. of Computer Engineering, Middle East Technical University (METU), Ankara, Turkey. S. Hicsonmez is  at the Dept. of Computer Engineering,Hacettepe University, Ankara, Turkey.
E-mail: { nermin@ceng.metu.edu.tr, samethicsonmez@hacettepe.edu.tr,  emre@ceng.metu.edu.tr }  \newline
\textsuperscript{\textdagger} Corresponding author.
}}

%
%

\markboth{Samet  \MakeLowercase{\textit{et al.}}: HoughNet: Integrating near and long-range evidence for visual detection}%
{Samet \MakeLowercase{\textit{et al.}}: Bare Demo of IEEEtran.cls for Computer Society Journals}
%



\IEEEtitleabstractindextext{%
\begin{abstract}
This paper presents HoughNet, a one-stage, anchor-free, voting-based, bottom-up object detection method. Inspired by the Generalized Hough Transform, HoughNet determines the presence of an object at a certain location by the sum of the votes cast on that location. Votes are collected from both near and long-distance locations based on a log-polar vote field. Thanks to this voting mechanism, HoughNet is able to integrate both near and long-range, class-conditional evidence for visual recognition, thereby generalizing and enhancing current object detection methodology, which typically relies on only local evidence.  On the COCO dataset, HoughNet’s best model achieves $46.4$ $AP$ (and $65.1$ $AP_{50}$), performing on par with the state-of-the-art in bottom-up object detection and outperforming most  major one-stage and two-stage methods. We further validate the effectiveness of our proposal in other visual detection tasks, namely, video object detection, instance segmentation, 3D object detection and keypoint detection for human pose estimation, and an additional ``labels to photo’’ image generation task, where the  integration of our voting module consistently improves performance in all cases. 
Code is available at \url{https://github.com/nerminsamet/houghnet}.
\end{abstract}

}

\maketitle

\IEEEdisplaynontitleabstractindextext

%
\IEEEpeerreviewmaketitle


\IEEEraisesectionheading{\section{Introduction}\label{sec:introduction}}
Deep learning has brought on remarkable improvements in object detection. Performance on widely used benchmark datasets, as measured by mean average-precision (mAP), has at least doubled (from 0.33 mAP \cite{voc-release5} \cite{felzenszwalb2009object} to 0.80 mAP on PASCAL VOC~\cite{resnet}; and from 0.2 mAP~\cite{mscoco} to around 0.5 mAP on COCO~\cite{retinanet}) in comparison to the pre-deep-learning, shallow methods. Current state-of-the-art, deep learning based object detectors \cite{faster, ssd, yolo2, retinanet} predominantly follow a top-down approach where objects are detected holistically via rectangular region classification. This was not the case with the pre-deep-learning methods. The bottom-up approach was a major research focus as exemplified by the prominent voting-based (the Implicit Shape Model \cite{ism}) and part-based (the Deformable Parts Model \cite{dpm-pami}) methods. However, today, among deep learning based object detectors, the bottom-up approach has not been sufficiently explored with a few exceptions (e.g. CornerNet \cite{cornernet}, ExtremeNet \cite{extremenet}). 


\begin{figure}

\captionsetup[subfigure]{labelformat=empty}
\centering
\begin{subfigure}[b]{0.19\textwidth} \fbox{\includegraphics[width=1.0\textwidth]{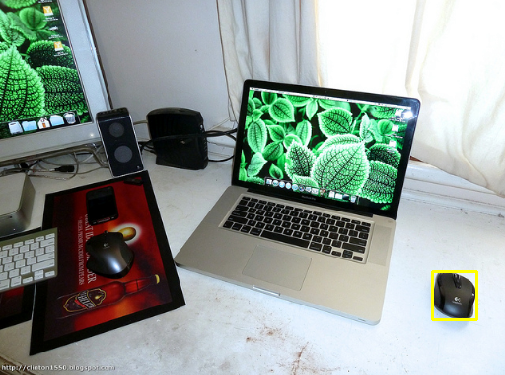}}
\end{subfigure} \hspace{2mm} 
\begin{subfigure}[b]{0.19\textwidth} \fbox{\includegraphics[width=1.0\textwidth]{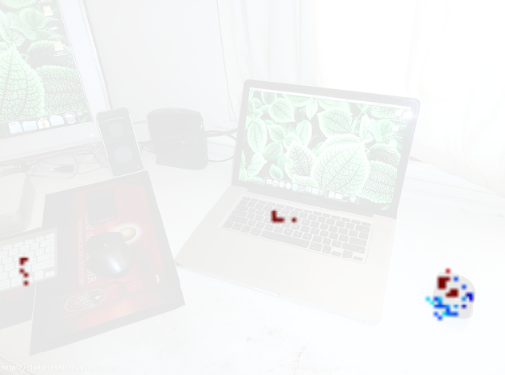}}
\end{subfigure} \hspace{2mm} 
\begin{subfigure}[b]{0.041\textwidth} \includegraphics[width=1.0\textwidth]{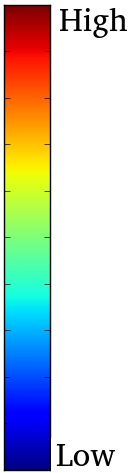}
\end{subfigure}

\caption{(Left) A sample ``mouse’’ detection, shown with yellow bounding box, by HoughNet. (Right) The locations that vote for this detection.  Colors indicate vote strength. In addition to the local votes originating from the mouse itself, there are strong votes from nearby ``keyboard’’ objects, which shows that HoughNet is able to utilize both short and long-range evidence for detection. More examples can be seen in Fig.~\ref{fig:vis-results}.}
\label{fig:teaser}
\end{figure}

In this paper, we propose HoughNet, a one-stage, anchor-free, voting-based, bottom-up object detection method. HoughNet is based on the idea of voting, inspired by the Generalized Hough Transform~\cite{hough1959,ght1981}. In its most generic form, the goal of GHT is to detect a whole shape based on its parts. Each part produces a hypothesis, i.e. casts its vote, regarding the location of the whole shape. Then, the location with the most votes is selected as the result. Similarly, in HoughNet,  the presence of an object belonging to a certain class at a particular location is determined by the sum of the class-conditional votes cast on that location (Fig.~\ref{fig:teaser}). HoughNet processes the input image using a convolutional neural network to produce an intermediate score map per class. Scores in these maps indicate the presence of visual structures that would support the detection of an object instance. These structures could be object parts, partial objects or patterns belonging to the same or other classes. We name these  score maps as ``visual evidence’’ maps. Each spatial location in a visual evidence map votes for target areas that are likely to contain objects. Target areas are determined by placing  a log-polar grid, which we call the ``vote field,’’ centered at the voter location. The purpose of using a log-polar vote field is to reduce the spatial precision of the vote as the distance between voter location and target area increases. This is inspired by foveated vision systems found in nature, where the spatial resolution rapidly decreases from the fovea towards the periphery~\cite{land2009looking}. Once all visual evidence is  processed through voting, the accumulated votes are recorded in object presence maps, where the peaks indicate the presence of object instances.

Current state-of-the-art object detectors rely on local or short-range visual evidence to decide whether there is an object at that location (as in top-down methods) or an important keypoint such as a corner (as in bottom-up methods). On the other hand, HoughNet is able to integrate both short and long-range visual evidence at the same time through voting. An example is illustrated in Fig.~\ref{fig:teaser}, where the detected mouse gets strong votes from two keyboards, one of which is literally at the other side of the image. In another example (Fig.~\ref{fig:vis-results}, row 2, col 1), a ball on the right-edge of the image is voting for the baseball bat on the left-edge.  On the COCO  dataset, HoughNet achieves comparable results with the state-of-the-art bottom-up detector CenterNet~\cite{centernet2}, while being the fastest among bottom-up detectors. It outperforms prominent one-stage (RetinaNet~\cite{retinanet}) and two-stage detectors (Faster R-CNN~\cite{faster}, Mask R-CNN~\cite{mask}).  To further show the effectiveness of our approach, we used the voting module of HoughNet in five other tasks, namely, video object detection, instance segmentation, 3D object detection, human pose estimation and ``labels to photo’’ image generation, and showed that the performances are consistently improved in all cases. 

To deal with the large number of training experiments  in model development and ablation studies, we curated ``COCO \texttt{minitrain}’’, a mini training set for COCO.   We validated COCO \texttt{minitrain} in two ways by (i) showing that the COCO \texttt{val2017} performance of a model trained on COCO \texttt{minitrain} is strongly positively correlated with the performance of the same model trained on COCO \texttt{train2017}, and (ii) showing that COCO \texttt{minitrain} set preserves the object instance statististics.

This paper extends our previous work \cite{samet2020houghnet} in several ways. (i) The original HoughNet~\cite{samet2020houghnet} applies voting only in the spatial domain (for object detection in still images). We extended this idea to the temporal domain by developing a new method, which takes the difference of features from two frames, and applies spatial and temporal voting using our ``temporal voting module’’ to detect objects. We showed the effectiveness of this  method in video object detection (Sections~\ref{sec:spatiotemporal} and \ref{sec:temporal-voting}). (ii) We showed that our Hough voting module is effective not only in object detection but also in other visual detection tasks as well: instance segmentation (Section~\ref{sec:instance-segmentation}), human pose estimation, i.e. keypoint detection (Section~\ref{sec:keypoint-det}) and 3D object detection (Section~\ref{sec:3d-object-det}). (iii) We added an analysis section (Section~\ref{sec:analysis}) where we examined the relations between vote-giver and vote-getter classes, and  evaluated HoughNet’s performance using localisation, recall, precision (LRP) metrics~\cite{lrp, new_lrp} to identify and compare the source of errors. (iv) We developed a ``scalable’’ variant of HoughNet where the number of voting operations does not depend on the number of object classes (Section~\ref{sec:scalable_model})  and showed that it dramatically improves inference speed with a slight drop in accuracy. (v) And finally, we improved the source code of HoughNet\footnote{\url{https://github.com/nerminsamet/houghnet}} by increasing its modularity and training speed.

Our main contribution in this work is HoughNet, a voting-based detection method that is able to integrate near and long-range evidence for visual detection tasks. 

\section{Related Work}

\textbf{Methods using log-polar fields/representations.} Many biological systems have foveated vision  where the spatial resolution decreases from the fovea (point of fixation) towards the periphery. Inspired by this phenomenon, computer vision researchers have used log-polar fields for many different purposes including shape description \cite{shapematching}, feature extraction  \cite{akbas2017object} and foveated sampling/imaging \cite{roboticlogpolar}.

\textbf{Non-deep, voting-based object detection methods.} In the pre-deep learning era, generalized Hough Transform (GHT) based voting methods have been used for object detection. The most influential  work  was  the Implicit Shape Model (ISM)~\cite{ism}. In ISM, Leibe \etal~\cite{ism} applied GHT for object detection/recognition and segmentation. During the training of the ISM, first, interest points are extracted and then a visual codebook (i.e. dictionary) is created using an unsupervised clustering algorithm applied on the patches extracted around interest points. Next, the algorithm matches the patches around each interest point to the visual word with the smallest distance. In the last step, the positions of the patches relative to the center of the object are associated with the corresponding visual words and stored in a table. During inference, patches extracted around interest points are matched to closest visual words. Each matched visual word casts votes for the object center. In the last stage, the location that has the most votes is identified, and object detection is performed using the patches that vote for this location. Later, ISM was further extended with discriminative frameworks~\cite{disc_hough, houghforest1, latent_hough, maxmargin, multiple_hough}. Okada~\cite{disc_hough} ensembled randomized trees using image patches as voting elements. Similarly, Gall and Lempitsky \cite{houghforest1} proposed to learn a mapping between image patches and votes using random forests. In order to fix the accumulation of inconsistent votes of ISM, Razavi~\etal~\cite{latent_hough} augmented the Hough space with latent variables to enforce consistency between votes.  In Max-margin Hough Transform~\cite{maxmargin}, Maji and Malik  showed the importance of learning visual words in a discriminative max-margin framework. Barinova~\etal~\cite{multiple_hough} detected multiple objects using energy optimization instead of non-maxima suppression peak  selection of ISM.

HoughNet is similar to ISM and its variants described above only at the idea level as all are voting based methods. There are two major differences: (i) HoughNet uses deep neural networks for part/feature (i.e. visual evidence) estimation, whereas ISM uses hand-crafted features; (ii) ISM uses a discrete set of visual words (obtained by unsupervised clustering) and each word’s vote is exactly known (stored in a table) after training. In HoughNet, however, there is not a discrete set of words and vote is carried through a log-polar vote field which takes into account the location precision as a function of target area.

\textbf{Bottom-up object detection methods.}  Apart from the classical one-stage~\cite{ssd, yolo3, dssd, rrc, retinanet} vs. two-stage~\cite{faster, mask} categorization of object detectors, we can also categorize the current approaches into two: top-down and bottom-up. In the top-down approach~\cite{ssd, yolo3, retinanet, faster},  a near-exhaustive list of object hypotheses in the form of rectangular boxes are generated and objects are predicted in a  holistic manner based on these boxes. Designing the hypotheses space (e.g. parameters of anchor boxes) is a problem by itself~\cite{region_proposals}. Typically, a single template is responsible for the detection of the whole object. In this sense, recent anchor-free methods \cite{ fcos, centernet} are also top-down. On the other hand, in the bottom-up approach, objects \emph{emerge} from the detection of parts or sub-object structures. For example, in CornerNet~\cite{cornernet}, top-left and bottom-right corners of objects are detected first, and then, they are paired to form whole objects. Following CornerNet, ExtremeNet~\cite{extremenet} groups extreme points  (e.g. left-most, etc.) and center points to form objects. Together with corner pairs of CornerNet~\cite{cornernet}, CenterNet~\cite{centernet2} adds center point to  model each object as a triplet. HoughNet follows the bottom-up approach based on a voting strategy: object presence score is voted (aggregated) from a wide area covering short and long-range evidence.

\textbf{Deep, voting-based object detection methods.}   Qi \etal~\cite{3dhough} apply Hough voting for 3D object detection in point clouds. Sheshkus \etal~\cite{otherhoughnet}  utilize Hough transform for vanishing points detection in the documents.  For automatic pedestrian and car detection, Gabriel \etal~\cite{hough_proposal} proposed using discriminative generalized Hough transform for proposal generation in edge images, later to further refine the boxes, they fed these proposals to deep networks.  In the deep learning era, we are not the first to use a log-polar vote field in a voting-based model. Lifshitz~\etal~\cite{consensus} used a log-polar map to estimate keypoints for single person human pose estimation. Apart from the fact that they are tackling a different problem (human pose estimation), there are several subtle differences. First, they prepare ground truth voting maps for each keypoint such that keypoints vote for every other one depending on its relative position in the log polar map. This requires manually creating static voting maps. Specifically, their model learns $H\times W\times R\times C$ voting map, where $R$ is the number of bins and $C$ is the augmented keypoints. In order to produce keypoint heatmaps,  they perform vote agregation at test phase. Second, this design restricts the model to learn only the keypoint locations as voters. When we consider the object detection task and its complexity, it is not trivial to decide on vote-giving locations for objects or prepare ground-truth voting maps as in human pose estimation. Moreover, this design limits the voters to reside only inside of the object (e.g. person), whereas in our approach an object could get votes from far away regions. To overcome these issues, unlike their model we apply vote aggregation during training (they perform vote agregation only at test phase). This allows us to expose the latent patterns between objects and voters for each class. In this way, our voting module is able to get votes from non-labeled objects  (e.g. ``candle’’, which is not a COCO class, voting for dining table; see the last row of Fig.~\ref{fig:vis-results}). To the best of our knowledge, we are the first to use a log-polar vote field in a voting-based deep learning  model to integrate the \textbf{long range interactions} for object detection.

Similar to HoughNet, Non-local neural networks (NLNN)~\cite{nlnn} and Relation networks (RN)~\cite{relation_networks} integrate long-range features.  As a fundamental difference, in NLNN, the relative displacement between interacting features is not taken into account. However, HoughNet uses this information encoded through the regions of the log-polar vote field. RN models object-object relations explicitly for proposal-based two-stage detectors.

Concurrently with HoughNet \cite{samet2020houghnet}, several object detectors have been introduced~\cite{detr, borderdet, relationnetplusplus}. Among them, the transformer-based DETR~\cite{detr}, in particular, shares with HoughNet the idea of using both short and long range interactions. In DETR, features at a specific location are updated through interactions with features  at other locations using encoder-decoder based transformers~\cite{attentiontransformers}. Although locations of features are taken into account using positional encoding, DETR models all possible pairs of features. Unlike DETR, HoughNet explicitly takes into account the spatial precision of the vote depending on the relative displacement between voter location and target (voted) area. Although attention-based neural networks and HoughNet share the idea of using long-range interactions, an in-depth comparison is beyond the scope of this work. Briefly; there are a variety of ways to encode location in attention-based neural networks: (i) simply no encoding -- location,  whether absolute or relative, is ignored \cite{nlnn}, (ii) using sine and cosine functions with varying frequencies \cite{detr, attentiontransformers}, (iii) using a general function \cite{koltunattentionexploring}. It is not trivial to compare these encodings to the explicit encoding of relative location using a log-polar field in HoughNet. Another fundamental difference is that transformers operate at the feature level by modifying features through interactions with other features, whereas HoughNet operates at the detection score level.  
\begin{figure*}
\includegraphics[width=\linewidth]{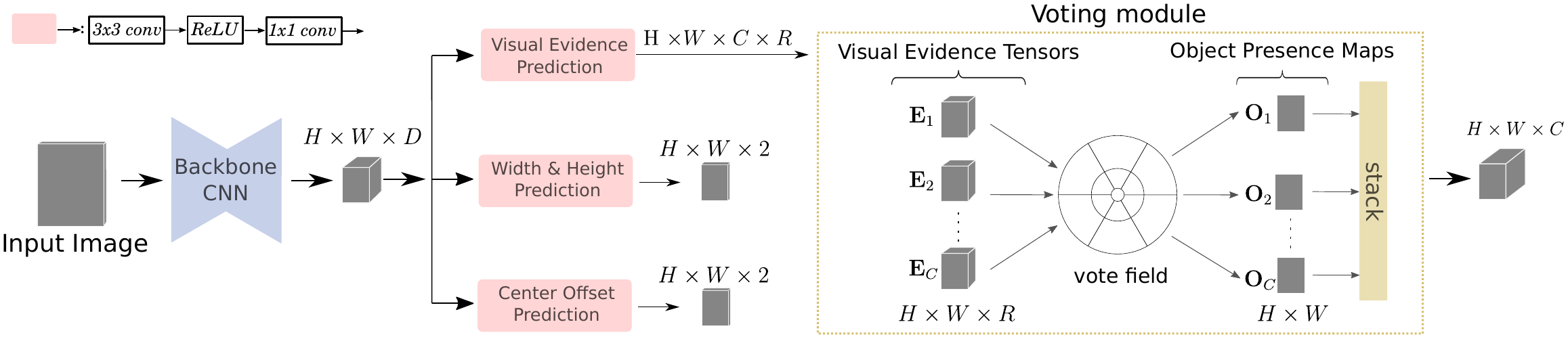}
\caption{Overview of the processing pipeline of HoughNet.}
\label{fig:model}
\end{figure*}

\section{HoughNet: models and method}

\textbf{Brief overview.} In HoughNet, the input image first passes through a backbone convolutional neural network (CNN), the output of which is connected to three different branches carrying out the predictions of (i) visual evidence scores, (ii) objects’ bounding box dimensions (width and height), and (iii) objects’ center location offsets. The first branch is where the voting occurs. Before we describe our voting mechanism in detail, we first introduce the log-polar vote field below. The overall processing pipeline of HoughNet is illustrated in Fig.~\ref{fig:model}.

\subsection{The Log-polar ``Vote Field’’}

We use the set of regions in a  standard log-polar coordinate system to define the regions through which votes are collected. A log-polar coordinate system is defined by the number and radii of eccentricity bins (or rings) and the number of angle bins. We call the set of cells or regions formed in such a coordinate system as the ``vote field’’ (Fig.~\ref{fig:vf}). In our experiments, we used  different vote fields with different parameters (number of angle bins, etc.) as explained in the Experiments section. In the following, $R$ denotes the number of regions in the vote field and $K_r$ is the number of pixels in a particular region $r$. $\Delta_r(i)$ denotes the relative spatial coordinates of the $i^{th}$ pixel in the $r^{th}$ region, with respect to the center of the field. We implement the vote field as a fixed-weight (non-learnable) transposed-convolution filter as further explained below.

\subsection{Voting Module}
\label{section:voting}

\begin{figure}
\includegraphics[width=0.4\linewidth]{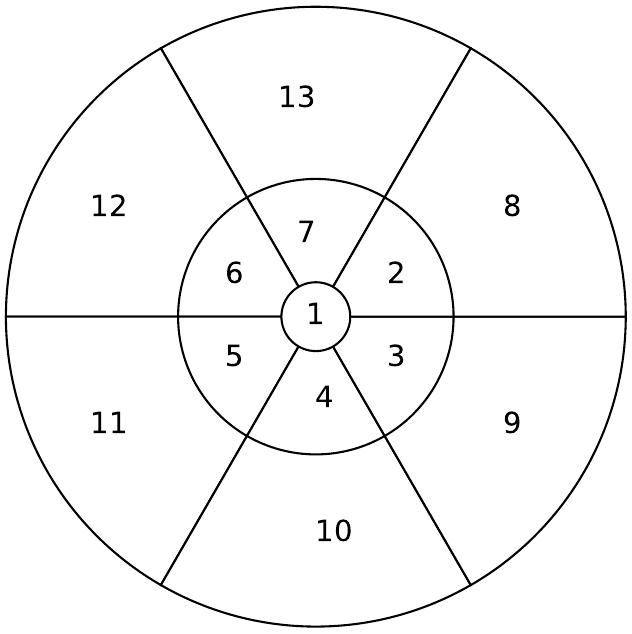}
\centering
\caption{A log-polar ``vote field’’ used in the voting module of HoughNet. Numbers indicate region ids. A vote field is parametrized by the number of angle bins, and the number and radii of eccentricity bins, or rings. In this particular vote field, there are a total of 13 regions, 6 angle bins and 3 rings. The radii of the rings are 2, 8 and 16, respectively.
}
\label{fig:vf}
\end{figure}

\begin{figure}
\includegraphics[width=.88\linewidth]{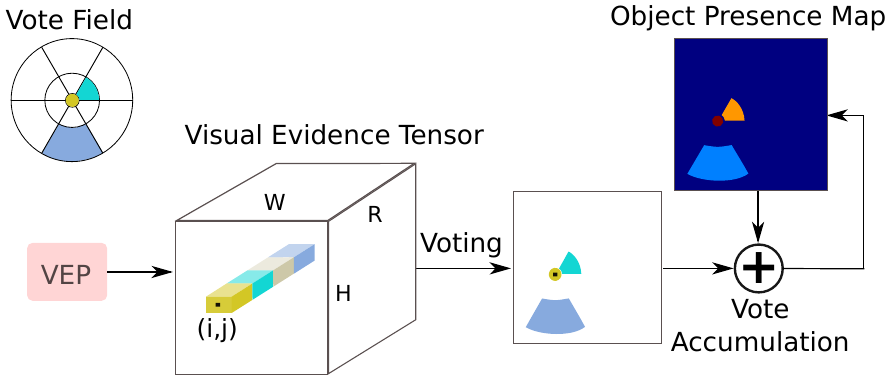}
\centering
\caption{Voting process for a single class. Visual evidence prediction (VEP) branch outputs $H\times W\times R$ dimensional visual evidence tensor for the class. Here the voting process is illustrated for just a single location $(i,j)$ and for 3 regions of the vote field shown with yellow, green and blue colors. The values at  $(i,j)$ corresponding to these 3 regions are added as votes to the appropriate locations in the Object Presence Map tensor. These locations are determined by the vote field shown at top left. Votes from all locations are similarly accumulated in the Object Presence Map. 
}
\label{fig:new_model_exp}
\end{figure}

After the  input image is passed through the backbone network and the ``visual evidence’’ branch, the voting module of HoughNet receives $C$ tensors $\mathbf{E}_1, \mathbf{E}_2, \dots, \mathbf{E}_C$, each of size  $H\times W\times R$, where $C$ is the number of classes, $H$ and $W$ are spatial dimensions and $R$ is the number of regions in the vote field. Each of these tensors contains class-conditional (i.e. for a specific class) ``visual evidence’’ scores. The job of the voting module is to produce $C$ ``object presence’’ maps $\mathbf{O}_1, \mathbf{O}_2, \dots, \mathbf{O}_C$, each of size $H\times W$. Then, peaks in these maps will indicate the presence of object instances. The voting process, which converts the visual evidence tensors (e.g. $\mathbf{E}_c$) to object presence maps (e.g. $\mathbf{O}_c$), works as described below.

The voting process is best explained using an example. Suppose we are  to process the visual evidence at the $i^\mathrm{th}$ row, $j^\mathrm{th}$ column and the $r^\mathrm{th}$ channel of a visual evidence tensor $\mathbf{E}$. We first place our vote field (Fig. \ref{fig:vf}) centered at $(i,j)$ on the $r^\mathrm{th}$ channel, which is a 2D map. The region $r$ of the vote field marks the target area to be voted on, whose coordinates can be calculated  by adding the coordinate offsets $\Delta_r(\cdot)$ to $(i,j)$. Then, we add the visual evidence score $\mathbf{E}(i,j,r)$ to the target area of the object presence map. Note that this operation can be efficiently implemented using the ``transposed convolution’’ (or ``deconvolution’’) operation. Visual evidence scores from locations other than $(i,j)$ are processed in the same way and the scores are accumulated in the object presence map. We formally define this procedure in Algorithm \ref{algo:voting}, which takes in a visual evidence tensor as input and produces an object presence map\footnote{We provide a step-by-step animation of the voting process in the supplementary material and at \url{https://shorturl.at/ilOP2}.}. Fig.~\ref{fig:new_model_exp} also  illustrates the defined voting process on a single class.

\begin{algorithm}
\caption{Voting process.}
\label{algo:voting}
\begin{algorithmic}
\REQUIRE Visual evidence tensor $\mathbf{E}_c$, Vote field relative coordinates $\Delta$
\ENSURE Object presence map $\mathbf{O}_c$

\STATE Initialize $\mathbf{O}_c$ with all zeros
\FOR{each pixel $(i,j,r)$ in $\mathbf{E}_c$}
\STATE /*  $K_r$: number of pixels in the vote field region $r$ */
\FOR{$k$ = 1 to $K_r$}
\STATE $(y,x) \leftarrow (i,j) + \Delta_r(k)$
\STATE $\mathbf{O}_c(y,x) \leftarrow \mathbf{O}_c(y,x) + \frac{1}{K_r} \mathbf{E}_c(i,j,r)$
\ENDFOR
\ENDFOR
\end{algorithmic}
\end{algorithm}

\begin{figure*}
\captionsetup[subfigure]{labelformat=empty}
\centering
\begin{subfigure}[b]{0.8\textwidth}  
\includegraphics[width=1.0\textwidth]{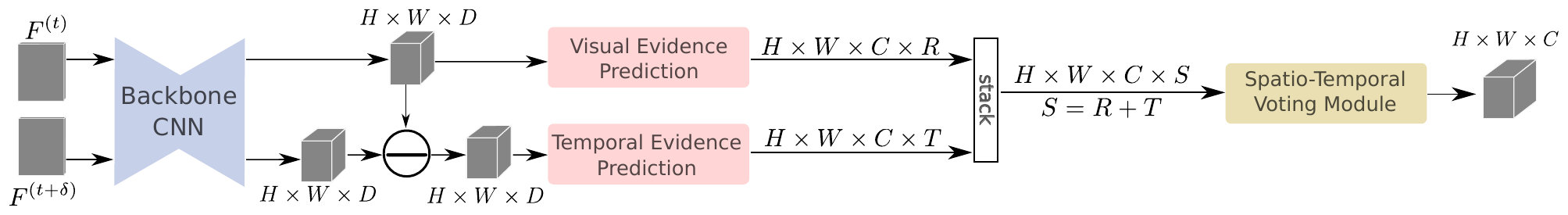}
\end{subfigure} \hspace{2mm} 
\begin{subfigure}[b]{0.1\textwidth} 
  \includegraphics[width=1.0\textwidth]{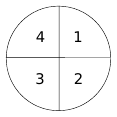} 
\end{subfigure} \hspace{2mm} 
 \caption{(Left) Overall processing pipeline of HoughNet for video object detection. (Right) Temporal vote field.}
\label{fig:temporal_model}
\end{figure*}

\subsection{Network Architecture}

Our network architecture design follows that of ``Objects as Points’’ (OAP)~\cite{centernet}. HoughNet consists of a backbone and three subsequent branches which predict  (i) visual evidence scores, (ii) bounding box widths and heights, and (iii) center offsets. Our voting module is attached to the visual evidence branch (Fig.~\ref{fig:model}).

The output of the backbone network is a feature map of size $H\times W\times D$, which is the result of an input image of size $4H\times 4W\times 3$. The backbone’s output is fed to all three branches.
Each branch has one convolutional layer with $3\times 3$  filters followed by a ReLU layer  and another convolutional layer with $1\times 1$ filters. The visual evidence branch outputs $H\times W\times C\times R$ sized output where $C$ and $R$ correspond to the number of classes and vote field regions, respectively.
The width \& height prediction branch outputs $H\times W\times 2$ sized output which predicts heights and widths for each object center.
Finally, the center offset branch predicts relative displacement of center locations across the spatial axes. These offsets help recover the lost precision of the center points due to down-sampling operations through the network. Both the width \& height branch and the center offset branch are class-agnostic.

\textbf{Objective functions.} For the optimization of the visual evidence branch, we use the modified focal loss~\cite{retinanet} introduced in CornerNet~\cite{cornernet} (also used in~\cite{extremenet, centernet}).  We optimize the center offset branch using the  $L_{1}$ loss as the other bottom-up detectors~\cite{cornernet, extremenet, centernet} do. Finally, for the width \& height prediction branch, we use  $L_{1}$ loss by scaling the loss by $0.1$ as proposed in OAP~\cite{centernet}. The overall loss is the sum of the losses from all branches.

\subsection{Spatio-temporal Voting}
\label{sec:spatiotemporal}
Unlike  static images, videos have rich temporal information. In order to benefit from the temporal clues in videos, researchers developed several methods to aggregate information locally and globally using  two or more frames~\cite{fgfa, ogemn, psla, mega}. Similarly, we extend HoughNet with a new temporal voting module to incorporate temporal information using an additional (auxiliary) frame.

Before describing the temporal voting process, we first explain the temporal log-polar vote field. The temporal vote field has 4 regions, $90^\circ$ angle bin and a single ring with radii 8. Each region of the temporal vote field stands for a motion direction. For example, the temporal vote-field region with id 1 in Fig.~\ref{fig:temporal_model}, corresponds to relative motion in $+x$ and $+y$ direction. When the temporal voting field is centered at a voter location, it votes for the locations in the target area depending on relative motion direction between frames.

We show the general processing pipeline for video object detection in Fig.~\ref{fig:temporal_model}. In addition to the reference frame at time $\textit{t}$, we choose a random auxiliary frame within $\delta$ time interval around the reference frame.   First, both frames pass through the backbone and we obtain feature maps of size $H\times W\times D$ for each. Then, we calculate motion features between these frames by subtracting the feature map of the auxiliary frame from the feature map of the reference frame. Bertasius~\etal  showed that using motion features is effective for pose detection in videos~\cite{dimoffs}. They interpret these features as discriminative motion features, where the network learns to ignore uninformative motion regions while focusing on discriminative motion cues.

Later, relative motion features are fed to the temporal evidence branch. Finally, visual and temporal evidence tensors are stacked and forwarded to the spatio-temporal voting module to output $H\times W\times C$ dimensional  object presence maps. Spatio-temporal voting aggregates votes in the same way as described in Section~\ref{section:voting}. Temporal evidence branch has the same architecture  as the visual evidence branch.

\section{Experiments}
\label{sec:exp}

In this section, we show the effectiveness of our proposed method on object detection and other visual detection tasks, namely, video object detection, instance segmentation, 3D object detection and keypoint detection (2D human pose estimation). Since we mostly focus on object detection, the object detection section is more detailed compared to others. It includes (i) ablation experiments through which we studied how different parameters of the vote field affect the performance, (ii) comparison with baseline, (iii) comparison with state-of-the-art, (iii) an analysis section and (iv) a scalable voting approach where the number of ``visual evidence tensors’’ does not depend on the number of object classes. The sections for other tasks mostly contain comparisons with baseline. We also include an image generation task in the context of ``label-to-photo’’ translation at the end. In order to handle a large volume of ablation experiments, we created a small training set for COCO, called ``COCO \texttt{minitrain}’’, which is described first in the following.  

We ran our experiments on 4 V100 GPUs. For training, we used $512\times512$ images unless stated otherwise. The training setup is not uniform across different experiments, mainly due to different backbones. However, the inference pipeline is common for all HoughNet models.  We extract center locations by applying a $3\times 3$ max pooling operation on object presence heatmaps and pick the highest scoring 100 points as detections. Then, we adjust these points using the predicted center offset values. Final bounding boxes are generated using the predicted width \& height values on these detections.

\subsection{COCO minitrain}

To facilitate model development and  faster analysis in ablation experiments, we carefully curated  a mini training set for COCO, dubbed ``COCO \texttt{minitrain}’’.  It is a subset of the COCO \texttt{train2017} dataset,  containing $25K$ images (about 20\% of \texttt{train2017}) and around $184K$ objects across $80$ object categories. We randomly sampled these images from the full set while preserving the following three quantities as much as possible: (i) proportion of object instances from each class, (ii) overall ratios of small, medium and large objects, (iii) per class ratios of small, medium and large objects.

\begin{figure}
\captionsetup[subfigure]{labelformat=empty}
\centering
\includegraphics[width=0.45\textwidth]{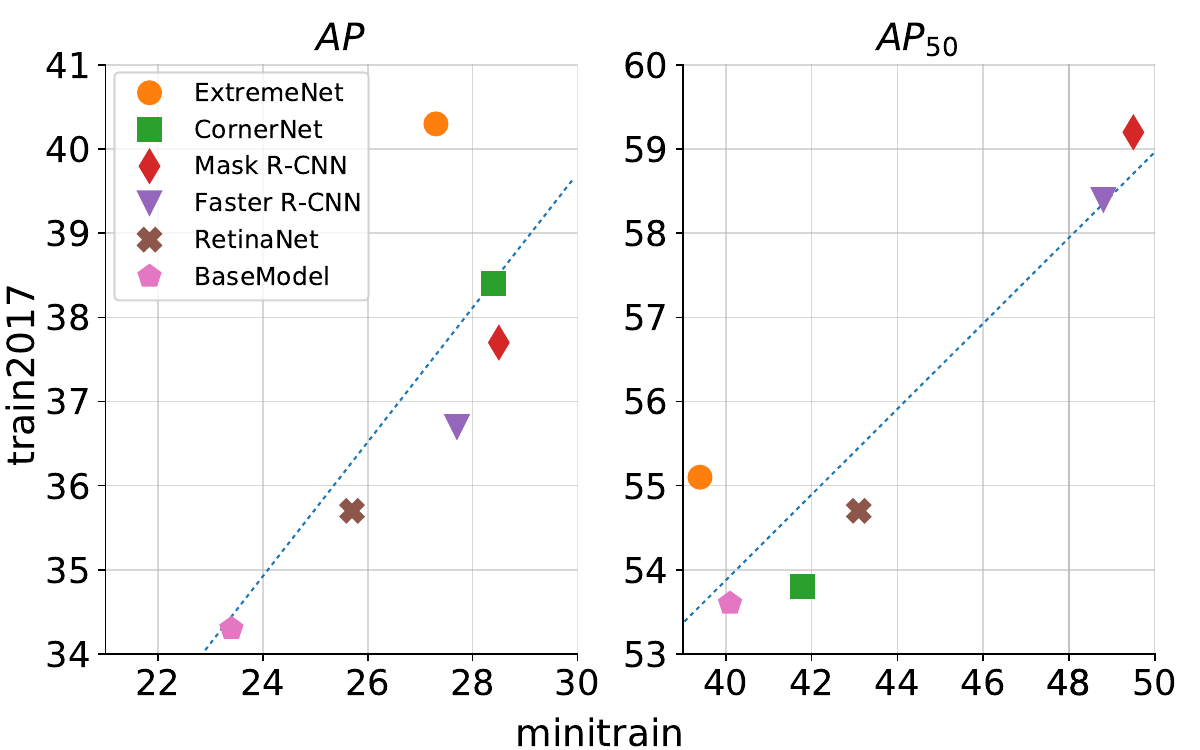} 
\caption{Performance validation for COCO \texttt{minitrain}. The x-axis is the \texttt{val2017} performance (AP on the left, AP50 on the right) of a model when it is trained on \texttt{minitrain}. The y-axis is the performance of the model when it is trained on the full COCO training set, \texttt{train2017}. Fitted lines with Pearson correlation coefficients $0.74$ and $0.92$, respectively for \textit{AP}  and \textit{AP$_{50}$},  show strong positive correlation.
}
\label{fig:minicoco}
\end{figure}

To validate COCO \texttt{minitrain}, we computed the correlation between the \texttt{val2017} performance of a model when it is trained on \texttt{minitrain} with the same as when it is trained on \texttt{train2017}. Over  six different object detectors (Faster R-CNN, Mask R-CNN, RetinaNet, CornerNet, ExtremeNet and HoughNet), the Pearson correlation coefficients turned out to be $0.74$  and $0.92$ for \textit{AP}  and \textit{AP$_{50}$},  respectively (Fig.~\ref{fig:minicoco}), which  indicate strong positive correlation. Further details on  \texttt{minitrain} and the dataset itself can be found at \url{https://github.com/giddyyupp/coco-minitrain}.

\subsection{Hough Voting for Object Detection}

In the  evaluation of object detection models, we follow the other bottom-up methods~\cite{extremenet, centernet, cornernet} and use two modes: (i) single-scale, horizontal-flip testing (SS testing mode), and (ii) multi-scale, horizontal-flip testing (MS testing mode). In MS, we use the following scale values, $0.6,1.0, 1.2, 1.5, 1.8$. To merge augmented test results, we use Soft-NMS~\cite{softnms}, and keep the top 100 detections. All tests are performed on a single V100 GPU.

\subsubsection{Ablation Experiments}
\label{sec:vot_abl}
Here we analyze the effects  of the number of angle and ring bins of the vote field on performance. Models are trained on COCO \texttt{minitrain} and evaluated on \texttt{val2017} set with SS testing mode. The backbone is Resnet-101~\cite{resnet}. In order to get higher resolution feature maps, we add three deconvolution layers on top of the default Resnet-101 network, similar to~\cite{xiao2018simple}. We add $3\times3$ convolution filters before each  $4\times4$ deconvolution layer, and put batchnorm and ReLU layers after convolution and deconvolution filters.  We trained the network with a batch size of 44 for 140 epochs with Adam optimizer~\cite{adam}. Initial learning rate $1.75 \times 10^{-4}$ was divided by 10 at epochs 90 and 120.

\textbf{Angle bins.} We started with a large, $65$ by $65$, vote field with 5 rings. We set the radius of these rings from the most inner one to the most outer one as $2$, $8$, $16$, $32$ and $64$ pixels, respectively. 
We experimented with $60^\circ$, $90^\circ$, $180^\circ$ and $360^\circ$ bins. We do not split the center ring (i.e. region with id 1 in Fig.~\ref{fig:vf}) into further regions.  Results are presented in Table~\ref{table:abl_table}a. For the $180^\circ$ experiment, we divide the vote field horizontally. $90^\circ$ yields the best performance considering both  \textit{AP} and  \textit{AP$_{50}$}. We used this setting in the rest of the experiments.

\begin{table}[]
\caption{Ablation experiments for the vote field. (a) Effect of angle bins on performance. Vote field with 90$^\circ$ has the best performance (considering \textit{AP} and  \textit{AP$_{50}$}).
(b) Effect of central and peripheral regions. Here, the angle bin is 90$^\circ$ and the ring count is four. Disabling any of center or periphery hurts performance, cf. (a). (c) Effect of number of rings. Angle is 90$^\circ$  and vote field size is updated according to the radius of the last ring. Using 3 rings yields the best result. It is also the fastest model.}
\centering
\resizebox{1.0\columnwidth}{!}{
\begin{tabular}{lcccccccc}
\toprule 
Model & \textit{AP}& \textit{AP$_{50}$} &  \textit{AP$_{75}$}&  \textit{AP$_S$} & \textit{AP$_M$} & \textit{AP$_L$}  & FPS \\
\midrule 
60$^\circ$     & \textbf{24.6} & 41.3     & 25.0     & \textbf{8.2} & 27.7 & 36.2 & 3.4     \\
90$^\circ$     & \textbf{24.6}& \textbf{41.5} & 25.0     & \textbf{8.2} & 27.7 & 36.2 & \textbf{3.5} \\
180$^\circ$    & 24.5       & 41.1      & 24.8     & 8.1         & 27.7 & \textbf{36.3} & \textbf{3.5}  \\
360$^\circ$    & \textbf{24.6} & 41.1     & \textbf{25.1} & 8.0   & \textbf{27.8} & \textbf{36.3} & \textbf{3.5} \\
\midrule 
\multicolumn{8}{c}{(a) Varying the Number of  Angle Bins}\\
\bottomrule 
Only Center     & 23.8 & 39.5 & 24.5 & \textbf{7.9} & 26.8 & 34.7 & \textbf{3.5} \\
No Center     & \textbf{24.4} & \textbf{40.9} & \textbf{24.9} & 7.4 & \textbf{27.6} & \textbf{37.1} & 3.3 \\
Only Context &  23.6  & 39.7 & 24.2 & 7.4 & 26.4 & 35.9 & 3.4 \\
\midrule 
\multicolumn{8}{c}{(b) Effectiveness of Votes from Center or Periphery}\\
\bottomrule 
5 Rings    & 24.6 & \textbf{41.5} & 25.0 & 8.2 & 27.7 & 36.2 & 3.5 \\
4 Rings    & 24.5 & 41.1 & 25.3 & 8.2 & \textbf{27.8} & 36.1 &   7.8 \\
3 Rings    & \textbf{24.8} & 41.3 & \textbf{25.6} & \textbf{8.4} & 27.6 & \textbf{37.5} & \textbf{15.6}\\
\midrule 
\multicolumn{8}{c}{(c) Varying Ring Counts}\\
\bottomrule 
\end{tabular}}
\label{table:abl_table}
\end{table}

\textbf{Effects of center and periphery.} We conducted experiments to analyze the importance of votes coming from different rings of the vote field. Results are presented in Table~\ref{table:abl_table}b. In the \textit{Only Center} case, we only keep the center ring and disable the rest. In this way, we only aggregate votes from features of the object center directly, which corresponds to a traditional object detector where only local (short-range) evidence is used.  This experiment shows that votes from outer rings help  improve performance. For the \textit{No Center} case,  we only disable the  center ring. We observe that there is only $0.2 $ decrease in \textit{AP}. This suggests that the evidence  for successful detection is embedded mostly around the object center not directly inside the object center. In order to observe the power of long-range votes, we conducted another experiment called ``Only Context,’’ where we disabled the two most inner rings and used  only the three outer rings for vote aggregation. This model  reduced \textit{AP} by $1.0$ point compared to the full model.

\begin{table}
\caption{Comparing our voting module to an equivalent dilated convolution filter, in terms of number of parameters and the spatial filter size, on COCO \texttt{val2017} set. Models are trained on COCO \texttt{train2017} and results are presented on SS testing mode.}
\centering
\resizebox{1.0\columnwidth}{!}{
 \begin{tabular}{lccccccc}
   \toprule 
   \textbf{Method} & \textit{AP}& \textit{AP$_{50}$} &  \textit{AP$_{75}$} & \textit{AP$_S$} & \textit{AP$_M$} & \textit{AP$_L$} \\
   \midrule 
    Baseline  & 36.2 & 54.8 & 38.7 & 16.3 & 41.6 & 52.3     \\
    + Dilated Conv. & 36.6 & 56.1 & 39.2 & 16.7 & 42.0 & 53.6     \\
    + Voting Module     & \textbf{37.3} & \textbf{56.6}    & \textbf{39.9}    & \textbf{16.8}    &
\textbf{42.6}    & \textbf{55.2}    \\
  \bottomrule
 \end{tabular}}
\label{table:dilated_compare}
\end{table}

\begin{table*}
 \caption{HoughNet results on COCO \texttt{val2017} set for different training setups. $^\dagger$ indicates initialization with CornerNet weights, $^*$~indicates initialization with ExtremeNet weights. Results are given for SS and MS testing modes, respectively.}
\centering
 \begin{tabular}{lcccccccc}
   \toprule 
   Models & Backbone & \textit{AP} & \textit{AP$_{50}$} &  \textit{AP$_{75}$} & \textit{AP$_S$} & \textit{AP$_M$} & \textit{AP$_L$} & FPS\\
   \midrule 
Base & R-101 & 36.0  \textbf{/} 40.7 & 55.2   \textbf{/} 60.6 & 38.4  \textbf{/} 43.9 & 16.2  \textbf{/} 22.5 & 41.7 \textbf{/} 44.2 & 52.0  \textbf{/} 55.7  & 3.5  \textbf{/} 0.5     \\


Base & R-101-DCN & 37.3 \textbf{/} 41.6 & 56.6   \textbf{/} 61.2 & 39.9   \textbf{/} 44.9 & 16.8   \textbf{/} 22.6 & 42.6  \textbf{/} 44.8 & 55.2   \textbf{/} 58.8 & 3.3 \textbf{/} 0.4     \\

Light & R-101-DCN & 37.2   \textbf{/} 41.5 & 56.5  \textbf{/} 61.5 & 39.6   \textbf{/} 44.5 & 16.8 \textbf{/} 22.5 & 42.5   \textbf{/} 44.8    & 54.9  \textbf{/} 58.4     &  14.3   \textbf{/} 2.1       \\
Light & HG-104
&    40.9  \textbf{/}   43.7 &  59.2  \textbf{/}   61.9    &  44.1   \textbf{/} 47.3 &  23.8 \textbf{/}  27.5 &  45.3   \textbf{/}  45.9 & 52.6   \textbf{/}  56.2 &  6.1   \textbf{/}  0.8   \\

Light & HG-104$^\dagger$
&    41.7  \textbf{/}  44.7 &  60.5  \textbf{/}  63.2    &  45.6   \textbf{/} 48.9 &  23.9 \textbf{/}  28.0 &  45.7   \textbf{/}  47.0 & 54.6   \textbf{/}  58.1 &  5.9   \textbf{/}  0.8    \\
Light & HG-104$^*$
&   43.0   \textbf{/}   \textbf{46.1} &  62.2  \textbf{/}   \textbf{64.6}    &  46.9  \textbf{/}   \textbf{50.3} &  25.5 \textbf{/}   \textbf{30.0} &  47.6   \textbf{/}   \textbf{48.8} & 55.8   \textbf{/}   \textbf{59.7} &  5.7   \textbf{/}  0.8    \\

  \bottomrule 
 \end{tabular}
\label{table:coco-compare}
\end{table*}

\begin{table*}
\caption{Comparison of object detection with baseline (OAP) on \texttt{val2017}. Results are given for SS and MS test modes, respectively.}
\centering
\begin{tabular}{ccccccccc}
  \toprule 
 {\textbf{Method} } &  \textit{AP}& \textit{AP$_{50}$} &  \textit{AP$_{75}$} & \textit{AP$_S$} & \textit{AP$_M$} & \textit{AP$_L$} & \textit{moLRP} $\downarrow$ \\
  \midrule 
  Baseline \textit{w} R-101-DCN  & 36.2  \textbf{/}  39.2 & 54.8  \textbf{/}  58.6  & 38.7   \textbf{/}  41.9 & 16.3  \textbf{/}  20.5& 41.6  \textbf{/}  42.6  & 52.3   \textbf{/}  56.2  &    71.1 \textbf{/}    68.3   \\
  + Voting Module     &  \textbf{37.2} \textbf{/} \textbf{41.5} & \textbf{56.5}   \textbf{/}\textbf{ 61.5} & \textbf{39.6}   \textbf{/} \textbf{44.5} & \textbf{16.8}   \textbf{/} \textbf{22.5} & \textbf{42.5}  \textbf{/} \textbf{44.8} & \textbf{54.9}   \textbf{/} \textbf{58.4}    &    \textbf{69.9} \textbf{/}    \textbf{66.6}  \\
  \midrule 
       Baseline  \textit{w} HG-104  & 42.2 \textbf{/}  45.1  & 61.1   \textbf{/}  63.5 & 46.0  \textbf{/}  49.3  & 25.2  \textbf{/} 27.8 & 46.4   \textbf{/}  47.7 & 55.2   \textbf{/}  \textbf{60.3}  &    66.1 \textbf{/}    63.9    \\
  + Voting Module     & \textbf{43.0}   \textbf{/}   \textbf{46.1 } &  \textbf{62.2}  \textbf{/}  \textbf{64.6} & \textbf{46.9}  \textbf{/}  \textbf{50.3} &  \textbf{25.5} \textbf{/}   \textbf{30.0} &  \textbf{47.6}   \textbf{/}   \textbf{48.8} & \textbf{55.8}   \textbf{/}   59.7  &     \textbf{65.6} \textbf{/}     \textbf{63.1}     \\
 \bottomrule
\end{tabular}
\label{table:baseline-comp}
\end{table*}

\textbf{Ring count.} To find out how far an object should get votes from, we discard outer ring layers one by one as presented in Table~\ref{table:abl_table}c. The models with $5$ rings, $4$ rings and $3$ rings have $17$, $13$ and $9$ voting regions and $65$, $33$ and $17$ vote field sizes, respectively. The model with $3$ rings yields the best performance on \textit{AP} metric and is the fastest one at the same time. On the other hand, the model with $5$ rings yields $0.2$ \textit{AP$_{50}$} improvement over the model with $3$ rings.

From all these ablation experiments, we decided to use the model with $5$ rings and $90^\circ$ as our \textit{Base Model}. Considering both speed and accuracy, we decided to use the model with $3$ rings and $90^\circ$ as our \textit{Light Model}.

\textbf{Voting module vs. dilated convolution.}
\label{sec:comp_abl}
Dilated convolution~\cite{dilated}, which can include long-range features, could be considered as an alternative to our voting module. To compare performance, we trained models on \texttt{train2017} and evaluated them on \texttt{val2017} using the SS testing mode.

\textit{Baseline}: We consider OAP with ResNet-101-DCN backbone as baseline. The last $1 \times 1$ convolution layer of center prediction branch in OAP, receives  $H \times W \times D$ tensor and outputs object center heatmaps with a tensor of size  $H \times W \times C$.

\textit{Baseline  +  Voting Module}: We first adapt the last layer of center prediction branch in baseline to output $H \times W \times C \times R$ tensor, then attach our voting module on top of the center prediction branch. Adding the voting module increases parameters of the layer by $R$ times. The log-polar vote field is $65 \times 65$, and has 5 rings ($90^\circ$). With 5 rings and $90^\circ$ we end up with $R=17$ regions.

\textit{Baseline + Dilated Convolution}: We use dilated convolution with kernel size $4 \times 4$ and dilation rate 22 for the last layer of the center prediction branch in baseline. Using  $4 \times 4$ kernel increases parameters  16 times which is approximately equal to $R$ in the \textit{Baseline  +  Voting Module}. Using dilation rate 22, the filter size becomes  $67 \times 67$ which is close to $65 \times 65$ log-polar vote field.


For a fair comparison with  \textit{Baseline}, both \textit{Baseline +  Voting Module} and the \textit{Baseline + Dilated Convolution} use ResNet-101-DCN backbone.
Our voting module outperforms dilated convolution in all cases (Table~\ref{table:dilated_compare}).

\subsubsection{Comparison with Baseline}
In Table~\ref{table:coco-compare}, we present the performance of HoughNet for different backbone networks, initializations and  our base-vs-light model, on the \texttt{val2017} set. There is a significant speed difference between Base and Light models. 
Our light model with R-101-DCN backbone is the second fastest one (14.3 FPS) achieving $37.2$ \textit{AP} and $56.5$ \textit{AP$_{50}$}. We observe that initializing the backbone with a pretrained model improves the detection performance.  In Table~\ref{table:baseline-comp}, we compare HoughNet’s performance with its baseline OAP~\cite{centernet} for two different backbones. HoughNet is especially effective for small objects, it improves the baseline by 2.1 and 2.2 \textit{AP} points for R-101-DCN and HG-104 backbones, respectively. We also provide results for the recently introduced \textit{moLRP}~\cite{lrp} metric, which combines localization, precision and recall in a single metric. Lower values are better.


\begin{table*}
\caption{Comparison with the state-of-the-art on COCO \texttt{test-dev}. The methods  are divided into three groups: two-stage, one-stage top-down and one-st
age bottom-up. The best results are boldfaced separately for each group.  Backbone names are shortened: R is ResNet, X is ResNeXt, F is FPN and HG is HourGlass.$^*$ indicates that the FPS values were obtained on the same AWS machine with a V100 GPU using the official repos in SS setup. The rest of the FPS are from their corresponding papers. F. R-CNN is Faster R-CNN.
}
\centering
\resizebox{1.0\textwidth}{!}{
\begin{tabular}{llcccccccccc}
  \toprule 
 Method & Backbone & Initialize & Train size & Test size &  \textit{AP}& \textit{AP$_{50}$} &  \textit{AP$_{75}$} & \textit{AP$_S$} & \textit{AP$_M$} & \textit{AP$_L$} & FPS\\
  \midrule 
  \textit{Two-stage detectors:} & & & & & & & & & & & \\
R-FCN~\cite{rfcn} & R-101 & ImageNet & 800$\times$800 & 600$\times$600 & 29.9 & 51.9 &  -  & 10.8 & 32.8 & 45.0 & 5.9 \\
CoupleNet~\cite{couplenet} & R-101 & ImageNet & ori. & ori. &  34.4 & 54.8 & 37.2 & 13.4 &  38.1 & 50.8 & -\\
F. R-CNN+++~\cite{resnet} & R-101 &  ImageNet &1000$\times$600 & 1000$\times$600& 34.9 & 55.7& 37.4& 15.6& 38.7 & 50.9 & - \\
F. R-CNN~\cite{fpn} & R-101-F & ImageNet &1000$\times$600 & 1000$\times$600& 36.2 & 59.1 & 39.0& 18.2 &39.0&48.2 & 5.0 \\
Mask R-CNN~\cite{mask} & X-101-F &  ImageNet & 1300$\times$800 & 1300$\times$800& 39.8 &  62.3 & 43.4 & 22.1 & 43.2 & 51.2 & 11.0 \\
Cascade R-CNN~\cite{cascade} & R-101 & ImageNet &  - & - & 42.8 & 62.1& 46.3 & 23.7& 45.5 &55.2& \textbf{12.0} \\
PANet~\cite{panet} & X-101 & ImageNet & 1400$\times$840 & 1400$\times$840& \textbf{47.4} & \textbf{67.2} & \textbf{51.8} & \textbf{30.1} & \textbf{51.7} & \textbf{60.0} & - \\
  \midrule 
\textit{One-stage detectors:} & & & & & & &  && &&\\
\textit{Top Down:} & & & & & & & && &&\\

SSD~\cite{ssd}     & VGG-16         &  ImageNet & 512$\times$512 & 512$\times$512 & 28.8 & 48.5 & 30.3 & 10.9 & 31.8 & 43.5     & -\\
YOLOv3~\cite{yolo3}    & Darknet          &  ImageNet & 608$\times$608 & 608$\times$608& 33.0 & 57.9 &  34.4 & 18.3 &  35.4 & 41.9     & \textbf{20.0}\\
DSSD513~\cite{dssd} & R-101     &  ImageNet & 513$\times$513 & 513$\times$513& 33.2& 53.3 & 35.2 & 13.0 & 35.4 &51.1    & -\\
RefineDet (SS)~\cite{refinedet} & R-101 & ImageNet & 512$\times$512 & 512$\times$512&  36.4 & 57.5 & 39.5 & 16.6 & 39.9 & 51.4 & - \\
RetinaNet~\cite{retinanet} & X-101-F & ImageNet & 1300$\times$800 & 1300$\times$800 & 40.8 & 61.1 & 44.1 & 24.1 & 44.2 & 51.2 & 5.4\\
RefineDet (MS)~\cite{refinedet} & R-101 & ImageNet & 512$\times$512 & $\leq$2.25$\times$ & 41.8 & 62.9 & 45.7 &  25.6 & 45.1 & 54.1 & - \\
OAP (SS)~\cite{centernet} & HG-104 & ExtremeNet &  512$\times$512 & ori. & 42.1 & 61.1 & 45.9 &  24.1 & 45.5 & 52.8 & 9.6$^*$ \\
FSAF (SS)~\cite{fsaf} & X-101 & ImageNet & 1300$\times$800 & 1300$\times$800 &   42.9  &  63.8 &  46.3 &  26.6 &  46.2  &  52.7 &  2.7 \\
FSAF (MS)~\cite{fsaf} & X-101 & ImageNet &  1300$\times$800 & $\sim\leq$2.0$\times$ &   44.6 &  65.2 &  48.6 &  29.7 &  47.1  &  54.6 &  - \\
FCOS~\cite{fcos}  & X-101-F  & ImageNet &  1300$\times$800 & 1300$\times$800 & 44.7 & 64.1 &  48.4 &  27.6 &  47.5 &    55.6 & 7.0$^*$ \\
FreeAnchor (SS)~\cite{freeanchor} & X-101-F & ImageNet &  1300$\times$960 & 1300$\times$960 & 44.9 & 64.3 & 48.5 & 26.8 & 48.3 & 55.9 & - \\
OAP (MS)~\cite{centernet} & HG-104 & ExtremeNet &  512$\times$512 & $\leq$1.5$\times$ & 45.1 & 63.9 & 49.3 &  26.6 & 47.1 & 57.7  & - \\
FreeAnchor (MS)~\cite{freeanchor} & X-101-F & ImageNet &   1300$\times$960 & $\sim\leq$2.0$\times$& \textbf{47.3} &   \textbf{66.3} &  \textbf{51.5} &   \textbf{30.6} & \textbf{50.4} &  \textbf{59.0} & - \\

\textit{Bottom Up:} & & & & & & & &&&& \\

ExtremeNet (SS)~\cite{extremenet} & HG-104 & - &  511$\times$511 & ori. & 40.2 & 55.5 & 43.2 & 20.4 & 43.2 & 53.1  & 3.0$^*$ \\
CornerNet (SS)~\cite{cornernet} & HG-104 & -  & 511$\times$511 & ori.& 40.5 & 56.5 & 43.1 & 19.4 & 42.7 & 53.9 & 5.2$^*$ \\
CornerNet  (MS)~\cite{cornernet}& HG-104 & -  & 511$\times$511 & $\leq$1.5$\times$ & 42.1 & 57.8 & 45.3 & 20.8 & 44.8 & 56.7 & - \\
ExtremeNet (MS)~\cite{extremenet} & HG-104 & - &  511$\times$511 & $\leq$1.5$\times$ & 43.7 & 60.5 & 47.0 & 24.1 & 46.9 & 57.6  & - \\
CenterNet (SS)~\cite{centernet2} & HG-104 & - &  511$\times$511 & ori. & 44.9 & 62.4 & 48.1 & 25.6 &  47.4 & 57.4 & 4.8$^*$ \\
CenterNet (MS)~\cite{centernet2} & HG-104 & - & 511$\times$511 & $\leq$1.8$\times$ &  \textbf{47.0} &  64.5 &  \textbf{50.7} & 28.9 &   \textbf{49.9} &  \textbf{58.9}  & - \\
\midrule 
HoughNet (SS)& HG-104 & -  & 512$\times$512 & ori. &  40.8 &   59.1 &   44.2 & 22.9 &  44.4 & 51.1 & \textbf{6.4}$^*$ \\
HoughNet (MS)& HG-104 & -  &  512$\times$512 & $\leq$1.8$\times$ & 44.0 &   62.4 &   47.7 & 26.4 &  45.4 & 55.2 & - \\
HoughNet (SS)& HG-104 &  ExtremeNet & 512$\times$512 & ori. &  43.1 & 62.2 & 46.8 & 24.6 &  47.0 & 54.4 & \textbf{6.4}$^*$  \\
HoughNet (MS)& HG-104 &  ExtremeNet &  512$\times$512 & $\leq$1.8$\times$ & 46.4 &  \textbf{65.1} &  \textbf{50.7} &  \textbf{29.1} &  48.5 &58.1 & -  \\
 \bottomrule 
\end{tabular}}
\label{table:stateoftheart-comp}
\end{table*}

\begin{figure*}[t]
\captionsetup[subfigure]{labelformat=empty}
\centering
\begin{subfigure}[b]{0.16\textwidth}
  \caption{Detection}
  \includegraphics[width=1.0\textwidth]{./figs/visual_res2/keyboard/15_det.jpg}
  \includegraphics[width=1.0\textwidth]{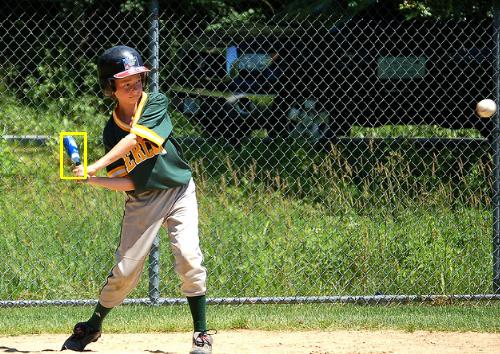}
  \includegraphics[width=1.0\textwidth]{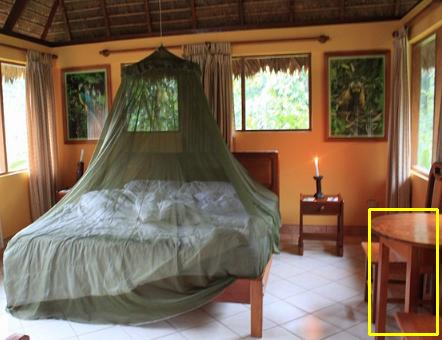}
\end{subfigure}
\begin{subfigure}[b]{0.16\textwidth}
  \caption{Voters}
  \includegraphics[width=1.0\textwidth]{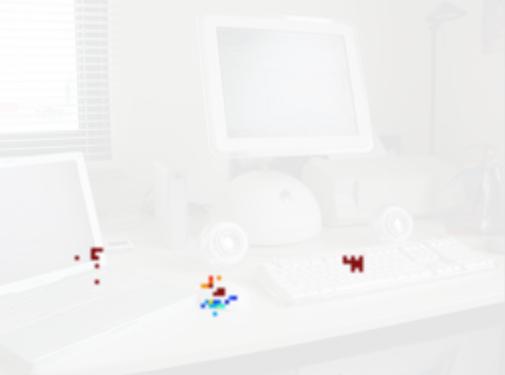}
\includegraphics[width=1.0\textwidth]{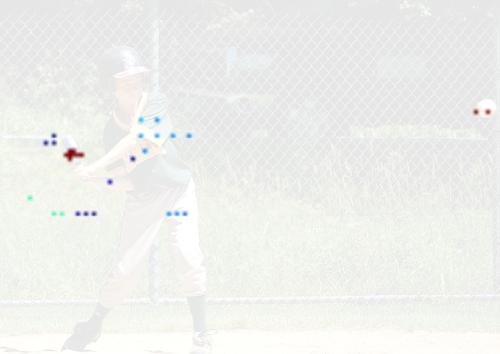}
  \includegraphics[width=1.0\textwidth]{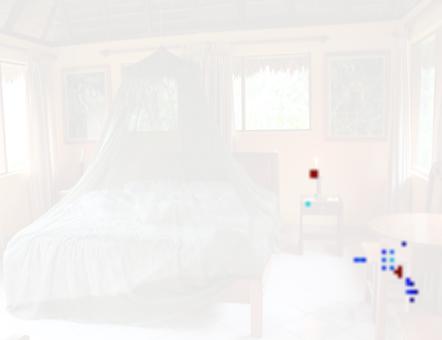}
\end{subfigure}
\begin{subfigure}[b]{0.16\textwidth}
  \caption{Detection}
  \includegraphics[width=1.0\textwidth]{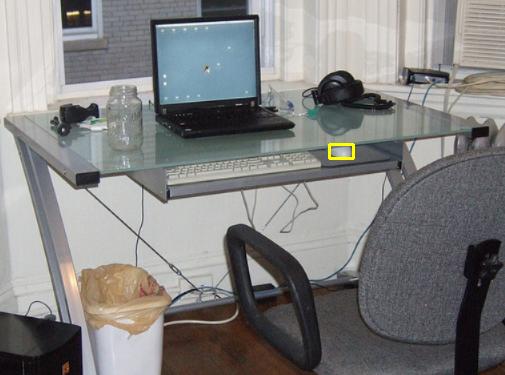}
  \includegraphics[width=1.0\textwidth]{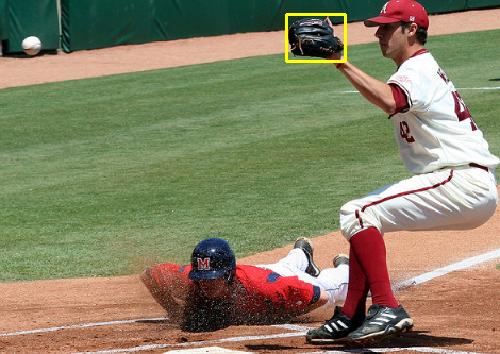}
  \includegraphics[width=1.0\textwidth]{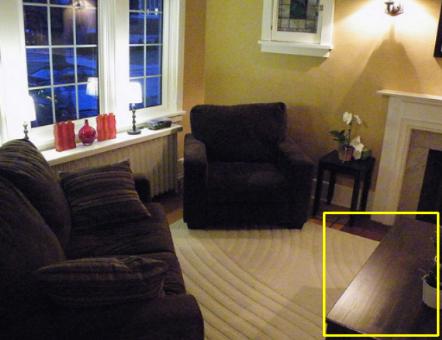}
\end{subfigure}
\begin{subfigure}[b]{0.16\textwidth}
  \caption{Voters}
  \includegraphics[width=1.0\textwidth]{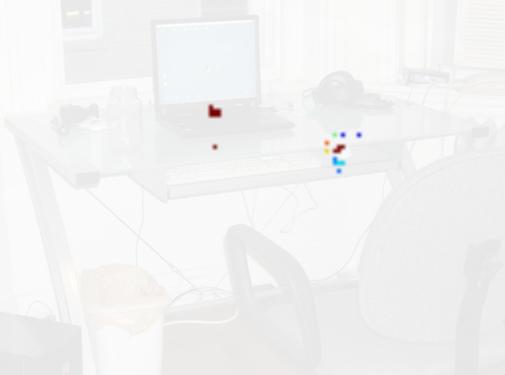}
\includegraphics[width=1.0\textwidth]{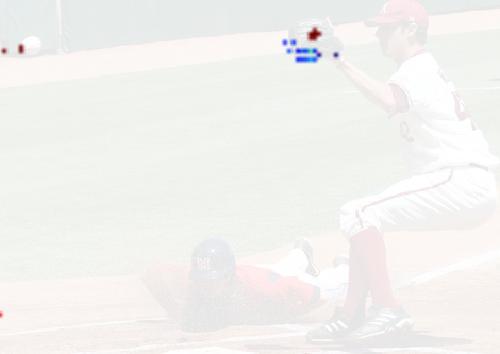}
 \includegraphics[width=1.0\textwidth]{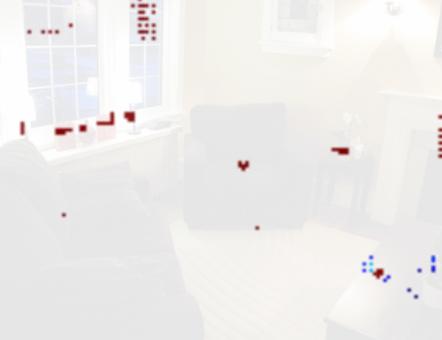}
\end{subfigure}
\begin{subfigure}[b]{0.16\textwidth}
  \caption{Detection}
  \includegraphics[width=1.0\textwidth]{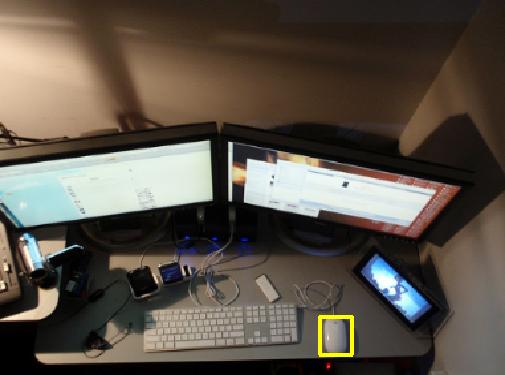}
  \includegraphics[width=1.0\textwidth]{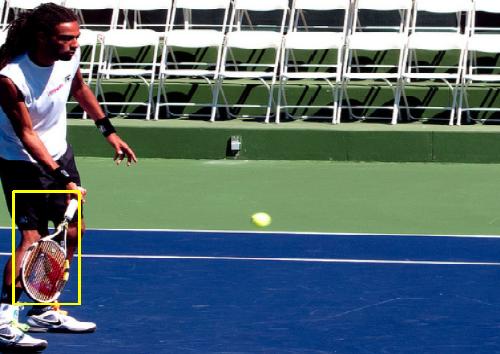}
 \includegraphics[width=1.0\textwidth]{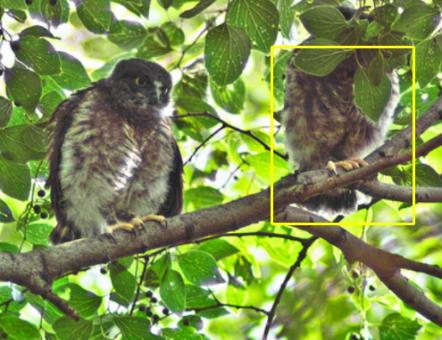}
\end{subfigure}
\begin{subfigure}[b]{0.16\textwidth}
  \caption{Voters}
  \includegraphics[width=1.0\textwidth]{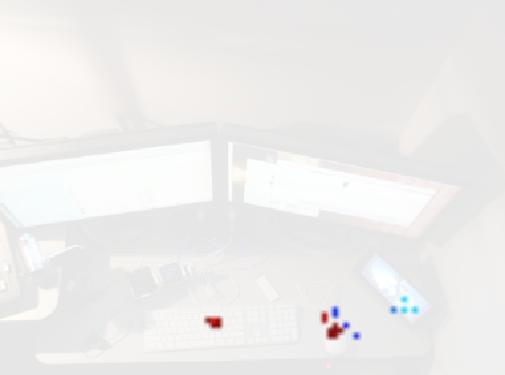}
\includegraphics[width=1.0\textwidth]{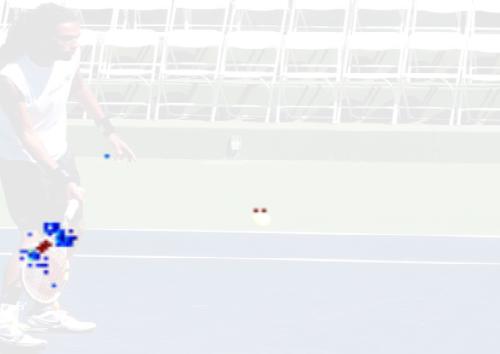}
 \includegraphics[width=1.0\textwidth]{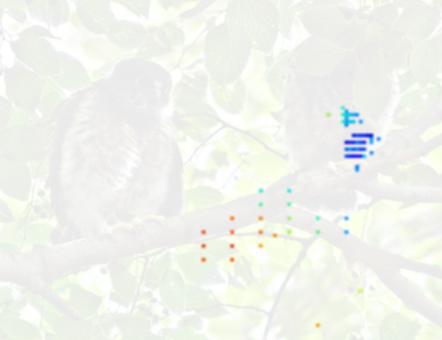}
\end{subfigure}
\caption{Sample detections of HoughNet and their vote maps. In the ``detection’’ columns, we show a correctly detected object, marked with a yellow bounding box. In the ``voters’’ columns, the locations that vote for the detection are shown. Colors indicate vote strength based on the standard ``jet’’ colormap (red is high, blue is low; Fig.~\ref{fig:teaser}).  In the \textbf{top row}, there are three ``mouse’’ detections. In all cases, in addition to the local votes (that are on the mouse itself), there are strong votes coming from nearby ``keyboard’’ objects. This voting pattern is justified given that mouse and keyboard objects frequently co-appear.  A similar behavior is observed in the detections of ``baseball bat’’, ``baseball glove’’ and ``tennis racket’’ in the \textbf{second row}, where they get strong votes from ``ball’’ objects that are far-away. In the first example of the \textbf{bottom row}, ``dining table’’ detection gets strong votes from the candle object, probably because they co-occur frequently. Candle is not among the 80 classes of COCO dataset. Similarly, in the second example in the  \textbf{bottom row}, ``dining table’’  has strong votes from objects and parts of a standard living room. In the last example, partially occluded bird gets strong votes (stronger than the local votes on the bird itself) from the tree branch.}
\label{fig:vis-results}
\end{figure*}

\begin{figure}
\captionsetup[subfigure]{labelformat=empty}
\centering
\includegraphics[width=0.42\textwidth]{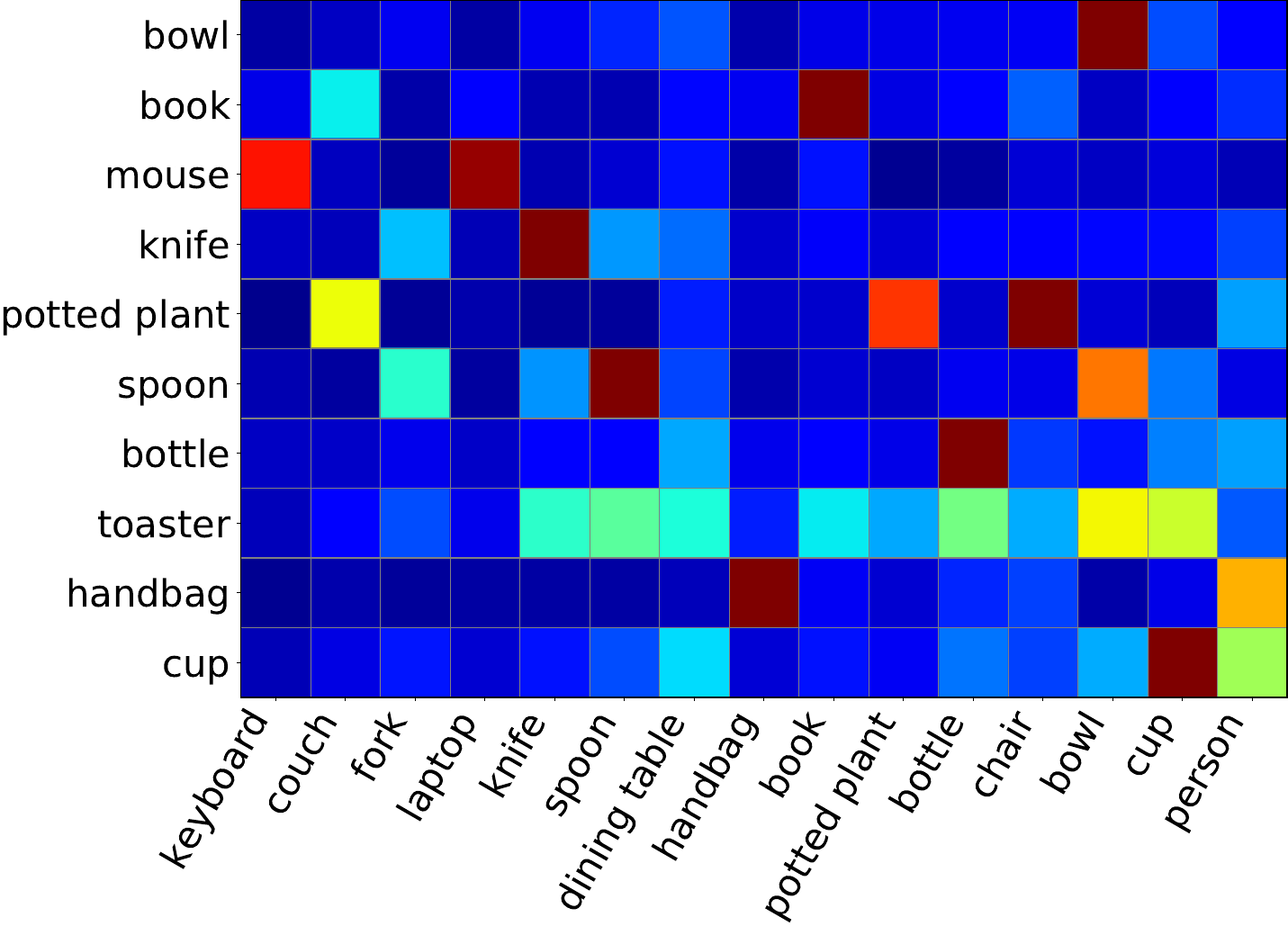}
\caption{ Object detection voting activity among 10 vote-getter classes (rows) and 15 vote-giver classes (columns) on the COCO dataset. Color decodes voting activity (red: high, blue: low). Each detection in a vote-getter class might get votes from multiple object classes. For example, the  \textit{cup} objects (i.e. the ``cup’’ row) get relatively high votes from \textit{cup}, \textit{person}, \textit{dining table} and \textit{bowl} classes. The full $80 \times 80$ matrix is provided in the supplementary material. 
}
\label{fig:vote-analysis}
\end{figure}

\subsubsection{Comparison with the State-of-the-art}
\label{sec:sota}

For comparison with the state-of-the-art, we use Hourglass-104~\cite{cornernet} backbone. We train Hourglass model with a batch size of 36 for 100 epochs using the Adam optimizer~\cite{adam}. We set the initial learning rate to  $2.5 \times 10^{-4}$ and divided  it by  10 at epoch 90.  Table~\ref{table:stateoftheart-comp} presents performances of HoughNet and several established state-of-the-art detectors. First,  we compare HoughNet with OAP~\cite{centernet} since it is the model on which we built HoughNet. In OAP, they did not present any results for “from scratch” training.  Instead they fine-tuned their model from ExtremeNet weights. When we do the same (i.e. initialize HoughNet with ExtremeNet weights), we obtain better results than OAP. However as expected, HoughNet is slower than OAP. Among the one-stage bottom-up object detectors, HoughNet performs on-par with the best bottom-up object detector by achieving $46.4$ \textit{AP} against $47.0$ \textit{AP} of CenterNet~\cite{centernet2}. HoughNet outperforms CenterNet on \textit{AP$_{50}$} ($65.1$ \textit{AP$_{50}$} vs. $64.5$ \textit{AP$_{50}$}). Note that, since our model is initialized with ExtremeNet weights, which makes use of the segmentation masks in its own training, our model  effectively uses more data compared to CenterNet. HoughNet is the fastest among one-stage bottom-up detectors. It is faster than CenterNet, CornerNet and more than twice as fast as ExtremeNet.



We provide visualization of votes for sample detections of HoughNet for qualitative visual inspection (Fig.~\ref{fig:vis-results}). These detections clearly show that HoughNet is able to make use of long-range visual evidence.

\begin{figure*}[h]
\includegraphics[width=0.8\linewidth]{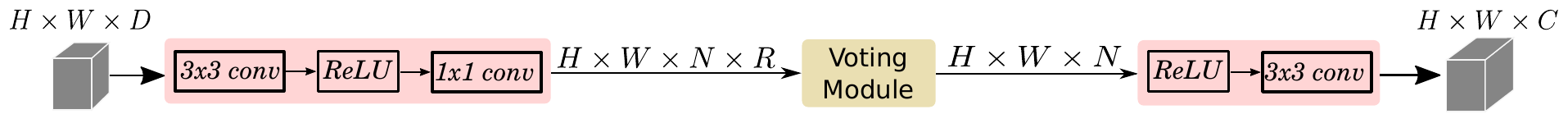}
\centering
\caption{A scalable variant of our voting process. Instead of having a separate visual evidence tensor for each of the $C$ classes, we create $N$ ($N << C$) visual evidence tensors that are shared and common to all classes.}
\label{fig:scalable_model}
\end{figure*}

\subsubsection{Analysis}
\label{sec:analysis}

Here we analyse the effect of Hough voting from two aspects. First, we inspect the error sources of both HoughNet and its baseline, and compare the two. This lets us understand how the voting module has improved the baseline. Secondly, we collect votes cast on a large number of images and analyse them to deduce interactions among object classes



\textbf{Error Sources.} 
Here we use the recently introduced ``localisation recall precision’’ (LRP) metric \cite{lrp, new_lrp}, which provides us with componentwise errors. These errors and the relative improvements yielded by the voting module are shown in Table \ref{table:error-baseline-comp}. The largest relative improvement ($+4.4\%$) is obtained in the ``false negative’’ (FN) component, which shows that the voting module increases the recall rate, that is, object instances normally missed by the baseline are successfully recovered with the integration of the voting module. This shows the importance of voting. ``False positive’’ (FP) component is also improved ($+3.3\%$), which indicates that object instances marked as background by the baseline are corrected by the voting module. These improvements come with no degradation in localisation performance.

\begin{table}
\caption{Effect of voting module on three major error types, namely, localisation (Loc), false positive (FP) and false negative (FN), as measured by the LRP metric \cite{lrp, new_lrp}. Relative improvement is calculated as $(B-V)/B$ where $B$ and $V$ are the LRP errors of baseline and  ``baseline+voting module”, respectively. Lower is better. Largest relative improvement is achieved in the FN component, which indicates that voting module increases the recall over the baseline.}
\centering
 
\begin{tabular}{cccc}
\toprule 
{\textbf{Model} } &  \textit{Loc } &  \textit{FP} & \textit{FN} \\
\midrule 
Baseline \textit{w} R-101-DCN   &  17.1  & 27.1 &  50.2 \\
+ Voting Module       &   17.1  & \textbf{26.2} &  \textbf{48.0} \\
\midrule 
Relative Improvement & 0\% & $+$3.3\% & $+$4.4\%  \\
\bottomrule
\end{tabular}
\label{table:error-baseline-comp}
\end{table}


\textbf{Interaction among object classes.}
We build a  $C \times C$ ($C$ is the number of classes) matrix to visualize voting relations among classes on the COCO dataset using the R-101-DCN backbone of HoughNet. In this matrix, rows represent vote-getters and columns represent vote-givers.  For each vote-getter class, we first  identify center points of detections and find all  locations that vote for these centers. Then, we sum class probabilities of all the voter-giver locations and obtain  a final $C$ dimensional vector that summarizes the votes received by the vote-getter class. Fig.~\ref{fig:vote-analysis} shows a $10 \times 15$ matrix, which we selected from the full $80 \times 80$ matrix based on the following criterion:   we selected the top 10 vote-getter classes with the maximum total votes. Next, for each vote-getter class, we selected the top 15 vote-giver classes that have the maximum voting activity. We provide the full $80 \times 80$ matrix in the supplementary material.

This matrix reflects many object co-occurrence relations and it captures object-to-object context well. For example,  \textit{mouse} class gets votes mostly from \textit{mouse}, \textit{keyboard} and \textit{laptop} classes. \textit{toaster} objects get votes from kitchen items. As a part of tableware items, both  \textit{spoon} and \textit{knife} get votes from each other and other tableware items like \textit{fork} and \textit{bowl}.  \textit{Potted plant} objects get from indoor objects such as \textit{chair} and \textit{couch}.


\subsubsection{A scalable approach to voting (independent of number of classes)}
\label{sec:scalable_model}

In HoughNet, there is a separate visual evidence tensor per object class (Fig. \ref{fig:model}) and a voting process is run for each of these  (Section \ref{section:voting}). This linear dependence on the number of object classes might be problematic when the number of classes increases dramatically, which is the case for newer datasets such as the 1000-class LVIS dataset \cite{lvis}.  With this in mind, we  designed a scalable variant of our voting module where,  instead of having a separate visual evidence tensor per class, we create $N$ tensors with $N << C$ (Fig. \ref{fig:scalable_model}). In this design, these $N$ tensors contain  “visual evidence” scores that are shared among and common to all classes. As a result, the voting process is carried out $N$ times instead of $C$ times. We convert the resulting $H\times W\times N$ dimensional voting maps to $H\times W\times C$ dimensional  object presence maps using a ReLU layer followed by a convolutional layer with $3 \times 3$ filters. We illustrate the scalable voting module in Fig.~\ref{fig:scalable_model}.

To analyze the scalable approach, we conducted ablation experiments for different values of $N$. We used the vote field from the \textit{Light Model} setup with $3$ rings and $90^\circ$ bins, using the ResNet-101-DCN backbone.  Models are trained on COCO \texttt{minitrain} and evaluated on \texttt{val2017} set in SS testing mode. Results are presented in Table~\ref{table:scalable_model_res}. The model with $N=8$ performs best among others.

To compare the scalable  model with the light model, we trained it on COCO \texttt{train2017} and obtained results on \texttt{val2017}. The scalable model is almost 2 times faster than the light model (26.3 FPS vs. 14.3 FPS) in SS testing mode.  However, it performs 1 \textit{AP} point worse than the base and light models with the same backbone, obtaining 36.3  \textit{AP}.

\begin{table}
\caption{Ablation experiments for the scalable voting module. Models are trained on COCO \texttt{minitrain} and results are presented on SS testing mode. Using $8$ visual evidence tensors yields the best result for all \textit{AP} metrics.}
\centering
\begin{tabular}{lccccccc}
 \toprule 
 \textbf{$N$} & \textit{AP}& \textit{AP$_{50}$} &  \textit{AP$_{75}$} & \textit{AP$_S$} & \textit{AP$_M$} & \textit{AP$_L$} \\
 \midrule 
  4  & 24.9 & 40.7 & 25.9 & 7.7 & 27.5 &  37.5   \\
  8 & \textbf{25.9} & \textbf{42.3}    & \textbf{26.7}    & \textbf{8.5}    &  \textbf{28.4}    &    \textbf{39.1}   \\
  12 & 24.6 & 40.6 & 25.6 & 7.8 & 27.0 & 36.6    \\
  16 & 25.0 & 41.1 & 25.8 & 8.2 & 27.4 & 37.9     \\   
\bottomrule
\end{tabular} 
\label{table:scalable_model_res}
\end{table}

\subsection{Hough Voting for Other Visual Detection Tasks}
\subsubsection{Video Object Detection}
\label{sec:temporal-voting}

We conducted our experiments on the ImageNet VID dataset~\cite{vid}. ImageNet VID dataset has 30 object categories. For training and evaluation, we followed widely adopted protocols~\cite{fgfa, mega} and trained our models on 3,862 videos in the training set and evaluated their performances on 555 videos in the validation set using mean average precision (mAP) as the evaluation metric. In addition to the overall mAP, we also present mAP results for slow, medium, and fast groups of videos as done in previous work~\cite{fgfa}. During the training of temporal models, we use a pair of video frames; a reference frame at time $\textit{t}$ and a nearby auxiliary frame at time $\textit{t} +  \delta$ where  $\delta$ is randomly picked from the set $\{-4,...,+4\}$.  At inference, we use fixed $\delta$ values and present our results  for $\delta=\{-4\}$ and $\delta=\{-4, +4\}$ settings.

First, we obtained single-frame baseline results for our baseline (OAP \cite{centernet}) and HoughNet. We trained OAP with 140 epochs using their official repository. For faster analysis, we trained our models with 80 epochs and divided the initial learning rate  $1.25 \times 10^{-4}$ by 10 at epoch 50. All models are trained with a batch size of 32.  Results in Table~\ref{table:vid_res} show that HoughNet outperforms its baseline (68.8 vs 65.0) in this setting.  Later, we conducted an ablation experiment to compare the performance of proposed temporal voting with feature aggregation over reference and auxiliary frames. To this end, both reference and auxiliary frames are passed through the backbone, then the output feature maps are concatenated  and fed to the visual evidence prediction branch. This model improved the single-frame baseline result of HoughNet by 2.7 mAP points. Next, we experimented with temporal voting using motion features. Temporal voting improved the performance further by 2.4 points. Using two auxiliary frames ($\delta$ is $\{-4,+4\}$), we obtained an even better result achieving 74.9 mAP.


\begin{table}
\caption{Results of video object detection on ImageNet VID validation set. Results are obtained without any test time augmentation. $\delta$ corresponds to the time offset of the auxiliary frame during inference.}
\centering
\resizebox{1.0\columnwidth}{!}{
\begin{tabular}{lccccc}
 \toprule 
{\textbf{Method} }  & $ \delta$
& \textit{mAP} &  \textit{mAP$_{F}$} & \textit{mAP$_{M}$} & \textit{mAP$_{S}$} \\
 \midrule 
\textit{Single-frame models:} & & & & & \\
 Baseline (OAP) & -  & 65.0  & 41.4 & 61.9 & 76.4    \\
 + Voting (HoughNet)   & -  &  68.8  & 45.8 & 66.1 & 79.1    \\
 \midrule
\textit{Temporal voting models} & & & & & \\
 +  Feat. Agg.  & {-4}   & 71.5  & 46.6 & 67.7 & 82.1   \\
+ Motion Features & {-4}   & 73.9  & 50.4 & 71.5 & 82.8   \\
+ Motion Features & {-4,+4}   & \textbf{74.9}  & \textbf{51.5} & \textbf{72.5} &\textbf{83.6}   \\
\bottomrule
\end{tabular}}
\label{table:vid_res}
\end{table}


\begin{table*}
\caption{Effect of voting module for the  instance segmentation task  on COCO \texttt{val2017} set. Results are shown for both COCO segmentation and box AP. Models are trained on COCO \texttt{train2017}. Results are obtained without any test time augmentation.}
\centering
 \begin{tabular}{lcccccc:cccccc}
   \toprule 
   \textbf{Method} & \textit{AP$^{seg}$}& \textit{AP$_{50}^{seg}$} &  \textit{AP$_{75}^{seg}$} &  \textit{AP$_S^{seg}$}  & \textit{AP$_M^{seg}$} & \textit{AP$_L^{seg}$} & \textit{AP$^{box}$}& \textit{AP$_{50}^{box}$} &  \textit{AP$_{75}^{box}$} &  \textit{AP$_S^{box}$} &  \textit{AP$_M^{box}$} & \textit{AP$_L^{box}$} \\
   \midrule 
    Baseline  & 27.2 &46.4 &28.0 & 8.6& 31.3& 43.9 & 33.9 & 51.3 & 36.5 &14.7 & 39.3 & 50.0  \\
    + Voting & \textbf{28.4} &  \textbf{48.0} &  \textbf{28.8} &  \textbf{9.1} &  \textbf{32.1} &  \textbf{46.1} &  \textbf{35.0}  &  \textbf{52.9}  &  \textbf{37.6} &  \textbf{15.0} &    \textbf{40.0} & \textbf{52.2}  \\
  \bottomrule
 \end{tabular}
\label{table:seg_results}
\end{table*}

\begin{table*}
\caption{Effect of voting module for the 3D object detection task on KITTI. Mean AP values of 5 models are presented with their standard deviations. Results are obtained without any test time augmentation.}
\centering
\begin{tabular}{cccc:ccc:ccc}
 \toprule 
{\textbf{Method} } &  \textit{AP$_{e}$}& \textit{AP$_{m}$} &  \textit{AP$_{h}$} & \textit{AOS$_e$} & \textit{AOS$_m$} & \textit{AOS$_h$} & \textit{BEV AP$_e$} & \textit{BEV AP$_m$} & \textit{BEV AP$_h$} \\
 \midrule 
 Baseline & \textbf{90.2}$\pm$1.2  &  80.4$\pm$1.4   & \textbf{71.1}$\pm$1.6   &  85.3$\pm$1.7 & 75.0$\pm$1.6   & 66.2$\pm$1.8  &  31.4$\pm$3.7 &  26.5$\pm$1.6 &  23.8$\pm$2.9   \\
 + Voting  & 89.2$\pm$0.4  &  \textbf{80.6}$\pm$3.0   & 70.0$\pm$0.2   &  \textbf{86.0}$\pm$0.8 & \textbf{77.0}$\pm$3.1   & \textbf{66.5}$\pm$0.6  &  \textbf{35.0}$\pm$0.8 &  \textbf{28.0}$\pm$0.8 &  \textbf{26.1}$\pm$0.7   \\
\bottomrule
\end{tabular}
\label{table:3dkitti_results}
\end{table*}

\subsubsection{Instance Segmentation}
\label{sec:instance-segmentation}

In order to show the effectiveness of voting for instance segmentation, we first extended our baseline OAP to perform instance segmentation. Inspired by the recent method BlendMask~\cite{blendmask}, we added to OAP  a new \textit{prototype mask prediction} branch to predict category independent, single prototype mask. This branch outputs a $H\times W\times 1$-dimensional feature map. In order to predict instance specific masks, 
we added another \textit{attention map prediction} branch which outputs $H\times W\times 196$ dimensional attention features.  These new branches have the same layer structure as other branches (see Fig.~\ref{fig:model}).

During training, we first get the \textit{instance specific local prototypes} for each instance by cropping them from the prototype mask according to their location and box size. Next, we extract $1\times 196$ \textit{instance specific attention maps} using the center points of instances, from the predicted attention map. Later we obtain $14\times 14$ instance specific attention maps by reshaping the attention features. We apply sigmoid normalization for both instance specific local prototypes and attention maps, and finally blend them by applying element-wise product. We use binary cross entropy as our loss function for segmentation.  To extend HoughNet for the instance segmentation task, we add the new branches as described above and follow the same training process. We provide the network architecture of instance segmentation for both baseline and HoughNet in the supplementary material.

We present the instance segmentation results on the COCO dataset in Table~\ref{table:seg_results}. Both baseline and HoughNet models are trained for 80 epochs with a batch size of 32. Initial learning rate $1.25 \times 10^{-4}$ is divided by 10 at epoch 50. Voting module of HoughNet improves instance segmentation performance by 1.2 AP  points and outperforms the baseline for all instance segmentation and box AP metrics.

\begin{figure}[t]
\captionsetup[subfigure]{labelformat=empty}
\centering
\begin{subfigure}[b]{0.2\textwidth}
 \caption{Detection}
 \includegraphics[width=1.0\textwidth]{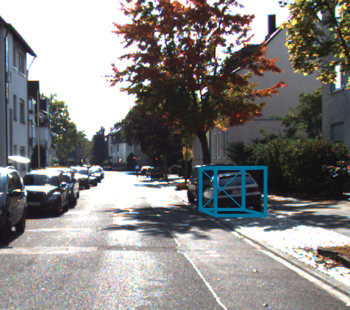}
\end{subfigure}
\begin{subfigure}[b]{0.2\textwidth}
 \caption{Voters}
 \includegraphics[width=1.0\textwidth]{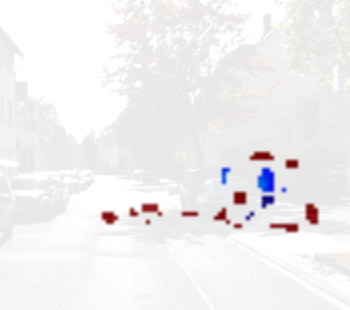}
\end{subfigure}
\caption{Sample \textit{car} detection of HoughNet from KITTI dataset and its vote map. We show a correctly detected object, marked with a blue 3D bounding box. In addition to the local votes, strong votes come from road and buildings.}
\label{fig:vis-results-3d}
\end{figure}

\textbf{Integrating Hough voting to Mask R-CNN:}  Mask R-CNN~\cite{mask} is an established state-of-the-art model for instance segmentation. It is an extension of the two-stage Faster R-CNN~\cite{faster}, where a new mask branch is added to predict object masks. Mask R-CNN~\cite{mask} consists of a backbone and a Feature Pyramid Network (FPN)~\cite{fpn}. Each FPN level outputs bounding boxes, class and mask predictions separately. Attaching our voting module at the end of each FPN level would be costly in terms of inference time. Moreover, it may require searching for the optimal parameters of the vote field for each FPN level. To overcome these issues, we follow Libra R-CNN~\cite{libra} which first resizes the multi-level FPN features to an intermediate size and then sums them to obtain a combined feature map. Libra R-CNN refines these combined features using a non-local neural network block \cite{nlnn}. In our experiments, we replace this non-local block with our voting module. Finally, refined features are rescaled and summed up with the output of each FPN level to strengthen the original features.
We conducted our experiments using ResNet-50~\cite{resnet} with FPN backbone. We use MMDetection~\cite{mmdetection} framework with Pytorch~\cite{pytorch} to implement our model. We present instance segmentation results in Table~\ref{table:maskrcnn-seg}. Our voting module improves detection and instance segmentation performances of Mask R-CNN by 1.1 AP and 1.2  AP$^{seg}$ points respectively, and outperforms the baseline for all AP metrics. 
\begin{table}
\centering
\caption{Effect of voting module when integrated to Mask R-CNN for the instance segmentation task on COCO \texttt{val2017} set. Results are shown for both COCO segmentation and box AP. Models are trained on COCO \texttt{train2017}. Baseline results are obtained using the publicly available Mask R-CNN model from MMDetection~\cite{mmdetection}.}
\begin{tabular}{lcccccc}
  \toprule 
  \textbf{Method} & \textit{AP} & \textit{AP$_{50}$}  & \textit{AP$_{75}$} & AP$^{seg}$ & AP$_{50}^{seg}$ & AP$_{75}^{seg}$ \\
  \midrule 
   Baseline  & 38.0 & 58.6 & 41.4 & 34.4 & 55.1 & 36.7  \\
   + Voting & \textbf{39.1}  &  \textbf{60.0}  & \textbf{42.5}  & \textbf{35.6} & \textbf{56.6} & \textbf{37.7}   \\
 \bottomrule 
\end{tabular}
\label{table:maskrcnn-seg}
\end{table}
\subsubsection{3D Object Detection}
\label{sec:3d-object-det}

Here we conduct experiments in 3D object detection to see whether our voting module is useful. In addition to the class and location prediction as done in 2D object detection,  a 3D object detector has to predict additional \textit{depth}, \textit{3D dimension} and \textit{orientation} attributes.  For this task also, we consider OAP~\cite{centernet} as our baseline. In addition to the branches from 2D object detection, OAP adds separate branches for these additional three attributes.

To show the effectiveness of the voting module, similar to 2D object detection, we attach our voting module with  $3$ rings and $90^\circ$ to the class prediction branch.  We experimented with \textit{car} classes of KITTI dataset~\cite{kitti} using the training and validation splits from SubCNN~\cite{sub}. Following OAP, we use the original image resolution $1280\times384$ both during training and inference.  We trained the network with a batch size of 16 for 70 epochs with Adam optimizer~\cite{adam}. Initial learning rate $1.25 \times 10^{-4}$ was divided by 10 at epochs 45 and 60. For fair comparison with the baseline, we use Deep Layer Aggregation (DLA)~\cite{dla} backbone as in OAP. More details related to both training and inference could be found in OAP~\cite{centernet}.

Table~\ref{table:3dkitti_results} shows results for bounding box AP, average orientation score (AOS) and bird-eye-view bounding box AP (BEV AP) at 3 levels of difficulty, namely, \textit{easy}, \textit{medium} and \textit{hard}. AP values in the table correspond to the average AP at IoU threshold 0.5  at 11 recalls from 0.0 to 1.0 with a step size of 0.1.  As also identified by OAP, since the recall threshold is small, the evaluation measures fluctuate up to 10\% \textit{AP}. To smooth the effect of fluctuations we trained 5 models and reported the averages together with standard deviation for each metric as in OAP.

The model with our voting module performs on par with the baseline for bounding box AP, and it outperforms baseline on AOS and BEV AP. Especially BEV AP is  significantly improved and  is more stable with much less variance compared to baseline. In Fig.~\ref{fig:vis-results-3d}, we show visualization of votes for sample \textit{car} detections.

\begin{figure}[t]
\captionsetup[subfigure]{labelformat=empty}
\centering
\begin{subfigure}[b]{0.15\textwidth}
 \caption{Detection}
 \includegraphics[width=1.0\textwidth]{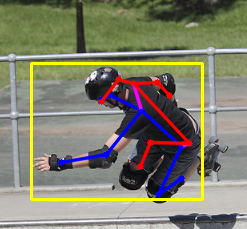}
\end{subfigure}
\begin{subfigure}[b]{0.15\textwidth}
 \caption{Voters}
 \includegraphics[width=1.0\textwidth]{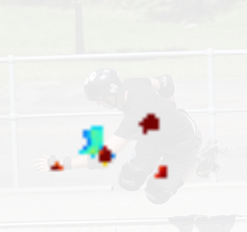}
\end{subfigure}
\begin{subfigure}[b]{0.15\textwidth}
 \caption{Voters}
 \includegraphics[width=1.0\textwidth]{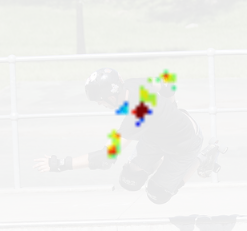}
\end{subfigure}
\caption{Sample keypoint detection of HoughNet and their vote maps. In the ``detection’’ column, we show a correctly detected object and its pose. Detection box is marked with a yellow bounding box, and pose estimation is shown with blue color for the left parts, red color for the right parts.  Votes for the \textit{left-elbow} and  \textit{left-shoulder} are shown respectively.  The \textit{left-elbow} detection gets strong votes from the \textit{left-shoulder}, \textit{left-knee} and \textit{left-wrist}.  The \textit{left-shoulder} detection gets votes from the area spanning from \textit{right-elbow} to \textit{left-elbow}. }
\label{fig:vis-results-kp}
\end{figure}

\begin{table*}
\caption{Comparing our voting module with baseline for 2D human pose estimation on COCO \texttt{val2017} set. The voting module is attached to the person classification and  keypoint estimation branches separately and concurrently. Results are shown for both COCO keypoint and box AP. Models are trained on COCO \texttt{train2017}. Results are presented on SS testing mode.}
\centering
\begin{tabular}{lccccc:cccccc}
  \toprule 
  \textbf{Method} & \textit{AP$^{kp}$}& \textit{AP$_{50}^{kp}$} &  \textit{AP$_{75}^{kp}$} & \textit{AP$_M^{kp}$} & \textit{AP$_L^{kp}$} & \textit{AP$^{box}$}& \textit{AP$_{50}^{box}$} &  \textit{AP$_{75}^{box}$} &  \textit{AP$_S^{box}$} &  \textit{AP$_M^{box}$} & \textit{AP$_L^{box}$} \\ 
  \midrule 
   Baseline  & 54.7 & \textbf{81.7} & 59.4 & 49.0 & 64.6 & 47.5 & 63.9 & 52.8 & 15.9 & 65.8 & 79.3    \\  
   + Voting for Person Class. & \textbf{56.9} & 81.6 & 61.9 & \textbf{51.3} & 67.9 &  50.1 & 71.4  & 54.1 &16.9 & 64.7 & 79.3   \\ 
   + Voting for Keypoint Est. & 56.8 & 81.5 & 61.2 & 50.7 & 68.0 & 50.2  & 70.9 & 54.3 & \textbf{17.2} & 64.4 & 79.4  \\ 
   + Voting for Both & \textbf{56.9} & 81.6 & \textbf{62.1} & 50.8 & \textbf{68.4} & \textbf{50.4}  & \textbf{71.7} & \textbf{54.4} & 17.0 & 64.9  & \textbf{79.8}   \\ 
 \bottomrule
\end{tabular}
\label{table:kp_results}
\end{table*}

\subsubsection{Keypoint detection: 2D Human Pose Estimation}
\label{sec:keypoint-det}

In this section, we show the effectiveness of our voting module for 2D Human Pose Estimation task. Human Pose Estimation is the problem of detecting human joints (i.e. keypoints) in images. It is another detection task that our voting module could help by leveraging the interactions both among keypoints and other visual parts.

Our baseline (OAP~\cite{centernet}) considers pose estimation as a regression task and defines each keypoint with an offset to the center of the instance and directly regresses them in a separate branch. In order to refine keypoints, OAP also employs the common bottom-up multi-human pose estimation approach as in~\cite{pose1,pose2,pose3} and predicts heatmaps for each of $k$ human joints in another branch. More detail on training and inference processes can be found in  OAP~\cite{centernet}.

We analyze the effect of voting by attaching our voting module  to the class prediction and the keypoint heatmap prediction branches both separately and at the same time. We conducted our experiments on COCO dataset which has $17$ keypoints for person object instances. For fair comparison with the baseline, we use DLA backbone and initialize it with the center detection model of OAP. We followed exactly the same training setup as in Section~\ref{sec:vot_abl}.  The models are trained on \texttt{train2017}  and evaluated on \texttt{val2017}. In Table~\ref{table:kp_results}, we show the results for both keypoint AP 
(\textit{AP$^{kp}$}) and box AP (\textit{AP$^{box}$}). Baseline results are obtained using the  publicly available equivalent model trained with 140 epochs from the official repository of OAP. In all cases, attaching our voting module improved the baseline. Using the  voting module in classification and keypoint heatmap prediction branches at the same time gives the best performance for both keypoint \textit{AP$^{kp}$} and box \textit{AP$^{box}$}. Attaching the voting module improves the baseline by $2.2$  and  $2.9$ points for \textit{AP$^{kp}$} and \textit{AP$^{box}$}, respectively. Especially in \textit{AP$_{50}^{box}$},  our voting module dramatically improved baseline by $7.8$ points. We provide visualization of votes for sample detections in Fig.~\ref{fig:vis-results-kp}

\begin{table}
\centering
\caption{Comparing our voting module and NLNN for object detection on OAP~\cite{centernet}. Models are trained on the COCO \texttt{train2017} set and results are presented on the COCO \texttt{val2017} set.}
\begin{tabular}{lccccc}
  \toprule 
  \textbf{Method} & \textit{AP} & \textit{AP$_{50}$}  & \textit{AP$_{75}$} & \textit{FPS} & \textit{Memory (GB)} \\
  \midrule 
   OAP + NLNN & 35.3 & \textbf{54.9} & 37.8 & 11 & 7.1  \\
   OAP + Voting & \textbf{35.7}  &  \textbf{54.9}  & \textbf{38.3}  & \textbf{19} & \textbf{1.5} \\
 \bottomrule 
\end{tabular}
\label{table:comparison-nlnn-centernet}
\end{table}

\subsection{Comparison with Non-local Neural Networks}

A fair comparison between our voting module and non-local neural networks (NLNN) is not trivial. Our voting module is intended to work with a classification head to aggregate class conditional evidence whereas non-local neural blocks are targeted to be integrated into backbones and necks (e.g. FPN~\cite{fpn}) to operate at lower resolutions. Therefore, for a fair comparison, we integrated both models into two different locations: (1) a classification head and (2) a detection neck, that is the Feature Pyramid Network (FPN) layer.

Our first set of experiments aim to  compare NLNN and our voting module when they are integrated to a classification head.  To this end, we add one non-local block right before the last layer of OAP’s~\cite{centernet} classification head and compare its results with HoughNet. Since the output of the classification head has higher spatial resolution compared to backbone features, NLNN’s memory usage and inference time increase significantly. In our experiments, we use ResNet-101-DCN backbone and present results  for single scale testing without any test time augmentations. As could be seen from Table~\ref{table:comparison-nlnn-centernet}, compared to NLNN, our voting module improves the baseline 0.4 AP points more, while being almost 2 times faster and consuming about 5 times less memory.

\begin{table}
\centering
\caption{Comparing our voting module and NLNN for object detection on Libra R-CNN. Models are trained on the COCO \texttt{train2017} set and results are presented on the COCO \texttt{val2017} set.}
\begin{tabular}{lccccc}
  \toprule 
  \textbf{Method} & \textit{AP} & \textit{AP$_{50}$}  & \textit{AP$_{75}$} & \textit{FPS} & \textit{Memory (GB)} \\
  \midrule 
   Baseline  & 36.8   & 58.0 & 40.0  & - & -  \\
   + NLNN & 37.7 & \textbf{59.4} & 40.9 & 11.3 & 2.0  \\
   + Voting & \textbf{37.8}  &  58.5    & \textbf{41.2}  & \textbf{11.5} & \textbf{1.8} \\
 \bottomrule 
\end{tabular}
\label{table:comparison-nlnn-libra}
\end{table}

Next, we  compare our voting module and non-local neural networks when both are integrated to a detection neck. Our comparison is conducted based on Libra R-CNN~\cite{libra}. Libra R-CNN by default adopts a non-local neural block for its feature refinement step at the end of the FPN. We replaced the non-local neural block in the feature refinement layer with our voting module. Experiments are conducted on the ResNet-50 backbone with FPN. We used Pytorch~\cite{pytorch} and MMDetection~\cite{mmdetection} for our implementation and followed the same training procedure as in Libra R-CNN. We present the results in Table~\ref{table:comparison-nlnn-libra}. The baseline model is the base Libra R-CNN model with IoU-balanced sampling where balanced L1 loss is ignored (see~\cite{libra} for the details). The model with our voting module performs on par with the Libra R-CNN model with a non-local block.
\subsection{Hough Voting for an Image Generation Task}
Another task where long-range interactions could be useful is the task of image generation from a given label map. There are two main approaches to solve this task; using unpaired and paired data for training. We take CycleGAN~\cite{cyclegan} and Pix2Pix~\cite{pix2pix} as our baselines for unpaired and paired approaches, respectively. We attach our voting module at the end of CycleGAN~\cite{cyclegan} and Pix2Pix~\cite{pix2pix} models.

\begin{figure}
\captionsetup[subfigure]{labelformat=empty}
\centering
\hfill
\begin{subfigure}[b]{0.14\textwidth}
 \caption{Input}
 \includegraphics[width=1.0\textwidth]{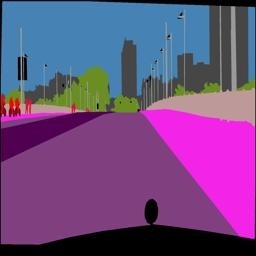}
\end{subfigure}
\begin{subfigure}[b]{0.14\textwidth}
 \caption{CycleGAN}
 \includegraphics[width=1.0\textwidth]{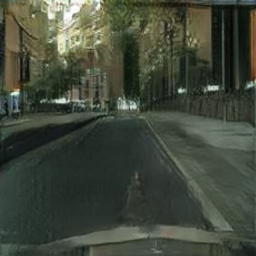}
\end{subfigure}
\begin{subfigure}[b]{0.14\textwidth}
 \caption{\shortstack{ + Voting}}
 \includegraphics[width=1.0\textwidth]{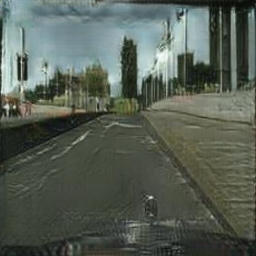}
\end{subfigure}
\hfill

~~~~
\caption{Sample qualitative results for the ``labels to photo’' task. When integrated with  CycleGAN, our voting module helps generate better images in the sense that the image conforms to the input label map better. CycleGAN fails to generate sky and buildings. For more visual results including Pix2Pix results please refer to~\cite{samet2020houghnet} or supplementary material.}
\label{fig:vis-results-gan}
\end{figure}

For quantitative comparison, we use the Cityscapes~\cite{cityscapes} dataset. In Table~\ref{table:gans}, we present FCN scores~\cite{fcn} (which is used as the measure of success in this task) of CycleGAN and Pix2Pix with and without our voting module. To obtain the ``without’’ result, we used the already trained model shared by the authors. We obtained the ``with” result using the official training code from their repositories. In both cases evaluation was done using the official test and evaluation scripts from their repos.  Results show that using the voting module improves FCN scores by large margins.
Qualitative inspection also shows that when our voting module is attached, the generated images conform to the given input segmentation maps better (Fig.~\ref{fig:vis-results-gan}). This is the main reason for the quantitative improvement. Since Pix2Pix is trained with paired data, generated images follow input segmentation maps, however, Pix2Pix fails to generate small details.

\begin{table}
\centering
\caption{Comparison of FCN Scores for the ``Labels to Photo’’ Task on the  Cityscapes Dataset~\cite{cityscapes}. Higher scores are better.}

\begin{tabular}{lccc}
  \toprule 
  \textbf{Method} & \textit{Per-pixel acc.}& \textit{ Per-class acc.} &  \textit{Class IOU } \\
  \midrule 
   CycleGAN  & 0.43   & 0.14 & 0.09       \\
   + Voting & \textbf{0.52} & \textbf{0.17} & \textbf{0.13} \\
   \midrule 
   pix2pix     & 0.71 &  \textbf{0.25} & 0.18 \\
   + Voting & \textbf{0.76}  &  \textbf{0.25}    & \textbf{0.20}  \\
 \bottomrule 
\end{tabular}
\label{table:gans}
\end{table}

\section{Conclusion}
In this paper, we presented HoughNet, a new, one-stage, anchor-free, voting-based, bottom-up object detection method.  HoughNet determines the presence of an object at a specific location by the sum of the votes cast on that location. Voting module of HoughNet is able to use both short and long-range evidence through its log-polar vote field. Thanks to this ability, HoughNet generalizes and enhances current object detection methodology, which typically relies on only local (short-range) evidence.  We show that HoughNet performs on-par with the state-of-the-art  bottom-up object detectors, and obtains comparable results with one-stage and two-stage methods. To further validate our proposal, we used the voting module of HoughNet in instance segmentation, 3D object detection, human pose estimation and ``labels to photo’’ image generation tasks, and showed that our voting module consistently improves the baseline  performances.


%



 

\ifCLASSOPTIONcompsoc
\section*{Acknowledgments}
\else
\section*{Acknowledgment}
\fi

This work was supported by the Scientific and Technological Research Council of Turkey (T\"{U}B\.{I}TAK) through the project titled ``Object Detection in Videos with Deep Neural Networks" (grant \#117E054). The numerical calculations reported in this paper were partially performed at T\"{U}B\.{I}TAK ULAKB\.{I}M, High Performance and Grid Computing Center (TRUBA resources). We also gratefully acknowledge the support of the AWS Cloud Credits for Research program.




{\bibliographystyle{IEEEtran}
\bibliography{references}
}

%

\begin{IEEEbiography}[{\includegraphics[width=1.0\textwidth]{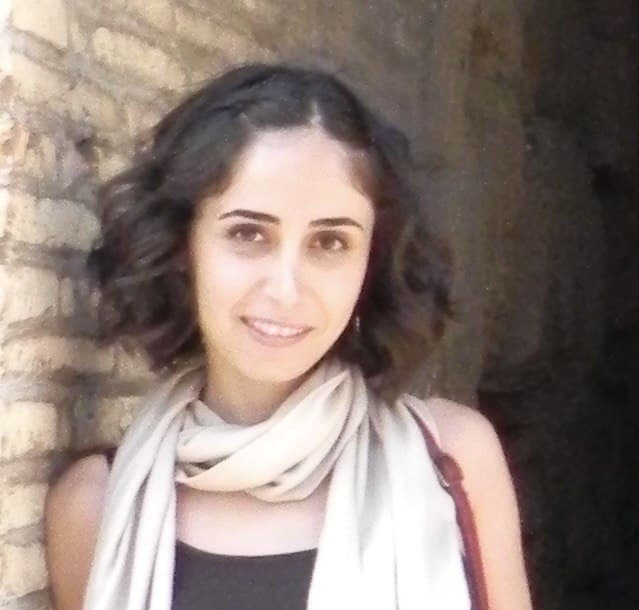}}]{Nermin Samet}
is a postdoctoral researcher at École des Ponts (ENPC) ParisTech, IMAGINE Research Group. She received her Ph.D. degree in 2021 from the Department of Computer Engineering, Middle East Technical University, Ankara, Turkey. Her  M.Sc. and B.Sc. degrees in computer engineering are from Bilkent University and Hacettepe University respectively. Her primary research interest is computer vision focusing on object detection.

\end{IEEEbiography}

\begin{IEEEbiography}[{\includegraphics[width=1.0\textwidth]{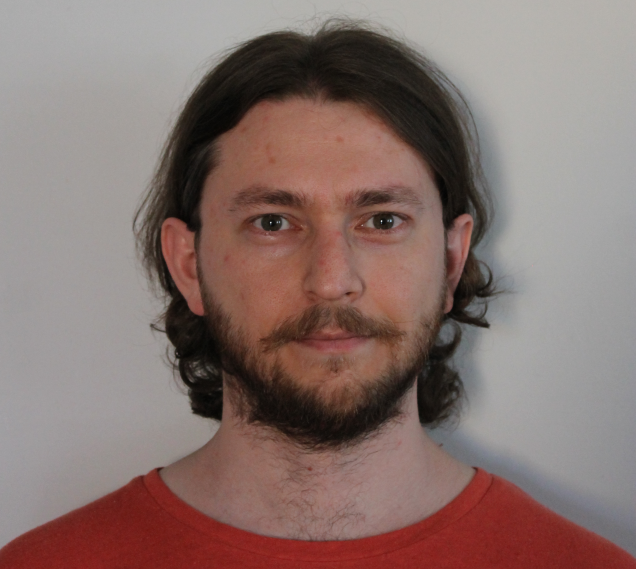}}]{Samet Hicsonmez}
is a Ph.D. candidate at the Department of Computer Engineering, Hacettepe University, Ankara, Turkey. Before joining Hacettepe, he received his M.Sc. degree from TOBB Economics and Technology University, Computer Engineering Department in 2013, and B.Sc. degree from Istanbul Technical University, Electronics Engineering in 2009. His primary research interests are computer vision focusing on GANs and object detection. 
\end{IEEEbiography}


\begin{IEEEbiography}[{\includegraphics[width=1.0\textwidth]{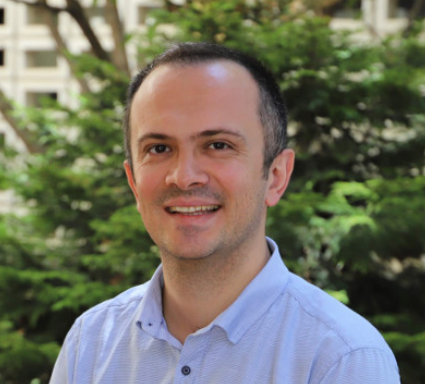}}]{Emre Akbas} 
is an assistant professor at the Department of Computer Engineering, Middle EastTechnical University (METU), Ankara, Turkey. He received his Ph.D. degree from the Departmentof Electrical and Computer Engineering, University of Illinois at Urbana-Champaign in 2011. His M.Sc. and B.Sc. degrees in computer science are both from METU. Prior to joining METU, he was a postdoctoral research associate at the Department of Psychological and Brain Sciences,University of California Santa Barbara. His research interests are in computer vision and deep learning with a focus on object detection and human pose estimation.
\end{IEEEbiography}




\end{document}